\documentclass[5p]{elsarticle}

\makeatletter
\def\ps@pprintTitle{%
 \let\@oddhead\@empty
 \let\@evenhead\@empty
 \def\@oddfoot{}%
 \let\@evenfoot\@oddfoot}
\makeatother

\usepackage{hyperref}

\usepackage{amsmath}
\usepackage{algpseudocode} 
\usepackage{graphicx}
\usepackage{amssymb}
\usepackage{algorithm}
\usepackage{paralist}
\usepackage{subfigure}
\usepackage[capitalize]{cleveref}
\usepackage{multirow}
\usepackage{makecell}
\usepackage[dvipsnames]{xcolor}
\usepackage{comment}
\usepackage{helvet} 
\usepackage[flushleft]{threeparttable}
\usepackage{booktabs,caption}
\usepackage{tikz}
\usetikzlibrary{positioning,automata}
\usepackage{forest,adjustbox}

\journal{Journal of Robotics and Autonomous Systems}

\begin{document}

\begin{frontmatter}

\title{\LARGE \bf Robot Gaining Accurate Pouring Skills through Self-Supervised Learning and Generalization}


\author{Yongqiang Huang}
\ead{yongqiang@usf.edu}

\author{Juan Wilches}
\ead{jwilches@usf.edu}

\author{Yu Sun\corref{mycorrespondingauthor}}
\cortext[mycorrespondingauthor]{Corresponding author}
\ead{yusun@usf.edu}

\address{Department of Computer Sci \& Eng, University of South Florida, 4202 E. Fowler Ave, Tampa, FL, United States, 33620}

\begin{abstract}
Pouring is one of the most commonly executed tasks in humans' daily lives, whose accuracy is affected by multiple factors, including the type of material to be poured and the geometry of the source and receiving containers. In this work, we propose a self-supervised learning approach that learns the pouring dynamics, pouring motion, and outcomes from unsupervised demonstrations for accurate pouring. The learned pouring model is then generalized by self-supervised practicing to different conditions such as using unaccustomed pouring cups. 
We have evaluated the proposed approach first with one container from the training set and four new but similar containers. The proposed approach achieved better pouring accuracy than a regular human with a similar pouring speed for all five cups. Both the accuracy and pouring speed outperform state-of-the-art works. We have also evaluated the proposed self-supervised generalization approach using unaccustomed containers that are far different from the ones in the training set.  The self-supervised generalization reduces the pouring error of the unaccustomed containers to the desired accuracy level.
\end{abstract}


\end{frontmatter}

\section{Introduction}

\begin{figure}
\centering
\includegraphics[width=0.95\columnwidth]{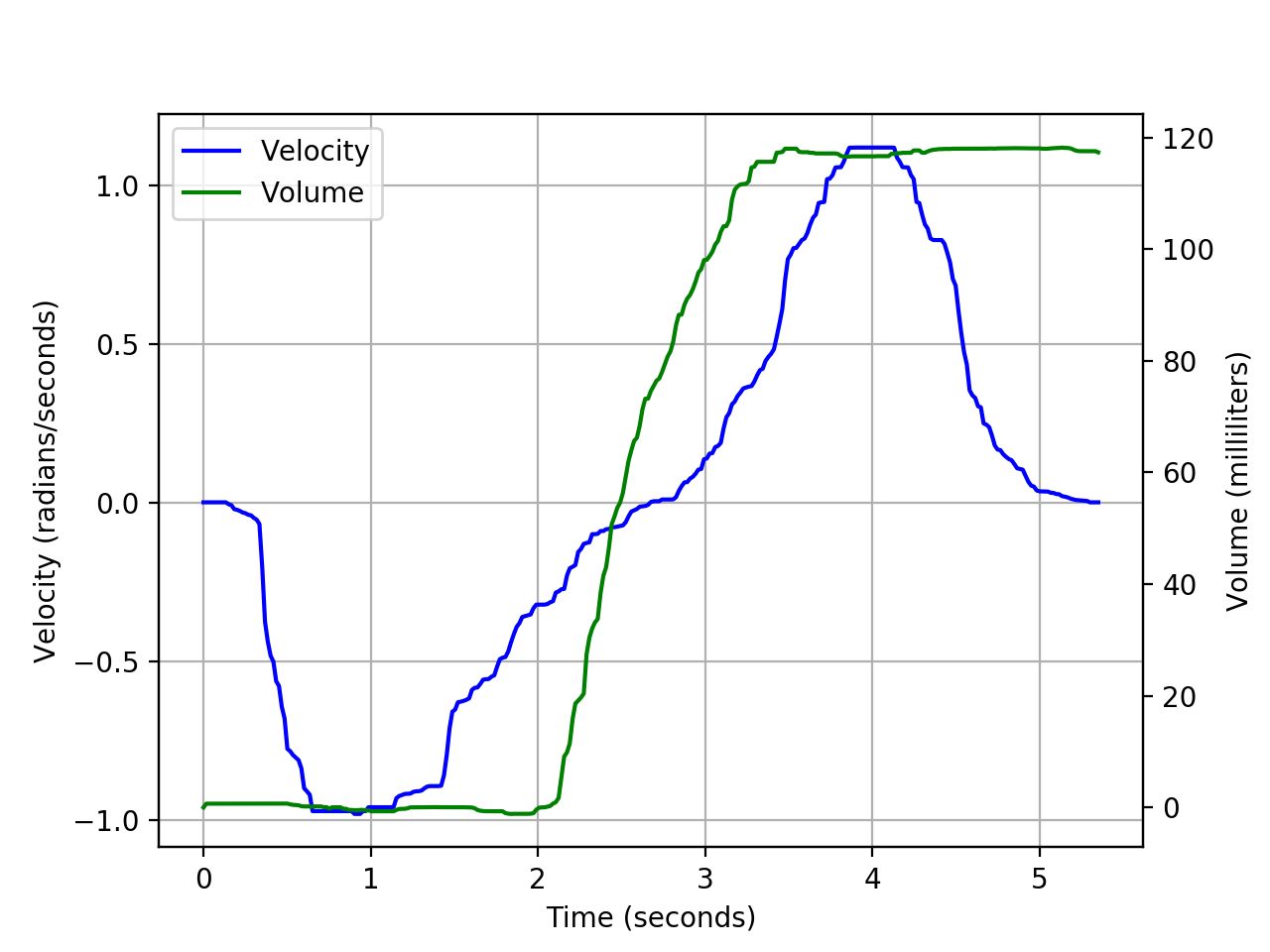}
\caption{An example of human pouring water. It can be seen that when the rotation is going backward, the volume is still increasing. Velocity: velocity of the source container measured counter clock-wise, negative radians/seconds means that the container is going forward and positive means backward. Volume: volume of water in the receiving container.}
\label{fig-human-pouring}
\end{figure}

Pouring is one of the most commonly executed tasks in humans' daily lives, especially for food preparation. In cooking scenarios, it is the most frequently executed motion \cite{pauliusiros2019}.  Pouring is a challenging task given the variety of containers and materials we can find in a kitchen. Humans excel at pouring liquids and solid materials, a skill that a cooking robot needs to master. This skill becomes particularly tricky when there is the need to execute it with precision and speed and for different setups and conditions. 
Accurate pouring is not a trivial task, and it is affected by many factors, including the property of the material, the geometry of the source and receiving containers, the manipulation of the source container, to name a few. It is a problem that cannot be solved using traditional control policies for two reasons:

\begin{enumerate}
    \item Lack of precise dynamics models: modeling fluid or granular motion precisely is either impossible or unfeasible because there are many unobservable parameters and those parameters vary with many factors such as the material and the shape of the pouring device.
    \item Un-reversible feature of the task: the poured material cannot come back to the pouring device once it is poured out. Therefore, it is not possible to have an overshoot in the system's response.   
\end{enumerate}
The two difficulties go hand-in-hand. The un-reversible feature of pouring calls for an approach that can predict. Figure \ref{fig-human-pouring} shows an example of velocity and volume sequences collected from a person pouring water. It can be seen that after the backward rotation starts, the water still comes out of the source container and the volume in the receiving container keeps increasing for a while. Therefore, the approach needs to predict when to start the backward rotation to reach the goal. However, prediction requires a precise model. In this paper, we propose a self-supervised learning approach that learns from demonstrations that are either unsupervised or performed by unskilled demonstrators.  The approach self-supervises the learning process by taking in all demonstrations without checking their performance or labeling them as successful or unsuccessful.  Instead, the self-supervised approach uses the real outcomes of the demonstrations as the desired goals. It is drastically different from traditional learning from demonstration approaches (LfD) \cite{Billard08chapter} that learn optimal motion trajectories from skilled demonstrators.  

We designed a data collection system that collected 284 human pouring demonstrations. This new data collection approach extends the Daily Interactive Manipulation (DIM) dataset \cite{huang2019dataset}. To learn water pouring dynamics, pouring motion, and outcomes, we have developed a peephole long short-term memory (LSTM) learning structure that used the previous step’s outcome as the current input.  The cell unit in the peephole LSTM learns, memorizes, and updates the liquid or granular material’s movement dynamics over time. It allows the peephole LSTM model to learn the relationship between the manipulation motion and the disparity between the current outcome and the desired outcome based on the liquid or granular material’s movement dynamics.

We have evaluated the proposed approach with five containers: one from the training set and four new containers that are not significantly different from the training set in terms of size. We have observed the mean volume error of those five containers to be as small as 4.12 milliliters (mL) and not higher than 12.35mL, a result that is better than the pouring accuracy of a regular person. The pouring speeds of the proposed approach are on par with a regular human pouring speed that is about 5 to 10 times faster than state-of-the-art approaches \cite{7989307, 8653969}. 
The learned model exhibits either much higher pouring error or much greater pouring standard deviation when evaluated on unaccustomed containers that are far different from the ones in the training set. To generalize the learned model to unaccustomed containers, we propose a self-supervised practicing approach. It uses the learned model to practice with the unaccustomed containers, collects the motions and outcomes from the practices, and uses the real outcomes as the desired outcomes to fine-tune the model.  We refer to this approach as {\it generalization by self-supervised practicing (GSSP)}, with which we achieved a reduction in mean volume error from values of more than 50\,mL to values lower than state-of-the-art works.  

Our contributions in this work include

\begin{enumerate}
\item Data collection system and pouring motion dataset. We designed a data collection system that captures the motion signals of human pouring employing a motion tracker and a force sensor (for volume measurement).  We have collected 284 pouring motions for nine different sizes of source containers.
\item Self-supervised learning from demonstrations and outcomes. We present a motion model for accurate pouring learned from human demonstrations. The pouring target in the input of the model during training is set to be the \textbf{actual poured volume} instead of the desired target. It allows the learning to be self-supervised since the approach does not have to specify or know the desired outcomes. 
\item The learned pouring skill could achieve human-like pouring accuracy and speed. The proposed approach models the complicated spatial-temporal patterns of human pouring, and as a result, pours smoothly and as fast as humans. In our dataset collected from human demonstrations, the pouring time ranges from 3.2 to 8.7 seconds, and the duration executed by our model ranges from 2.8 to 7.6 seconds. It pours faster than related methods \cite{7989307, 8653969} which report 25 seconds and 20-45 seconds per pour, respectively. It achieves lower pouring error than existing methods that also use a single modality to monitor the poured volume \cite{7989307, do2018accurate, chaudo2018, tianze2019}.   
\item Generalizing pouring skills by self-supervised practicing. We present the Generalization by self-supervised Practicing (GSSP) approach, which fine-tunes the model using the actual pouring outcomes of a robot. It allows the robot to pour accurately, using unaccustomed containers and materials. 

\end{enumerate}

\subsection{Related Work on LfD}
 
One popular LfD approach is Gaussian mixture regression (GMR) \cite{4650593}. The approach first learns the spatial-temporal relation of a motion. Then, it produces a novel sequence of the motion by producing the position or state of the motion corresponding to each time step. Functional Principal Component Analysis (fPCA) is another approach for spatial-temporal motion learning. The traditional PCA can be applied on both the temporal and spatial axes of a motion to find how the motion varies at different time steps \cite{lim2005, min2009}. Functional PCA (fPCA) extends PCA by representing a motion in a continuous-time format instead of using a collection of points \cite{ramsay_etal2009, dai2013functional, huang2015, paulius2016}.
The spatial-temporal GMR and fPCA in nature consider the motion as a whole rather than a dynamical system, which makes it inconvenient to receive timely feedback while executing the motion.
In comparison, GMR can be configured to learn the relationship between states and actions and thus behave as a dynamical system \cite{Hersch08TRO}.

Alternatively, one can use movement primitives (MP). The first MPs, the dynamic movement primitives (DMP), consist of three components: 1) a strictly damped string model that guarantees the convergence of the motion state to a goal state, 2) a forcing function that contains the shape that the motion is expected to go through, and 3) a canonical system that modulates the temporal profile of the motion \cite{Ijspeert:2013:DMP:2432779.2432781}. Its variants include but are not limited to interactive primitives (IP), which enable the interaction between two agents \cite{6907265} and probabilistic movement primitives (ProMP), which enable more flexibility on the force function \cite{NIPS2013_5177}.   
The general GMR, MP, and fPCA involve the usage of a temporal alignment algorithm of the motion data, such as dynamic time warping (DTW) \cite{sakoe_etal1978}, which may damage certain spatial-temporal patterns in the data in unclear ways. 
In comparison, a recurrent neural network (RNN) does not require aligning the motion data in time. It is designed to process time sequences and is capable of representing dynamical systems \cite{han2004, trischler2016}. RNN has been successfully applied for text generation as well as motion generation \cite{sutskever2014, graves2013, huang2017learning_pour}. 

\subsection{Related Work on Pouring} \label{sec:related_works}

In \cite{Pan:2016:RMP:3038594.3038659}, the authors propose trajectory planning algorithms for liquid body transfer that uses fluid simulation, \cite{Pan2017FeedbackMP} learns to predict the state of the fluid using neural networks. However, the fluid model of the liquid is in general difficult to obtain, for which reason \cite{TAMOSIUNAITE2011910} proposes adding goal learning to shape learning with MP for liquid transfer. \cite{6614613} considers the amount of the liquid in the source container while pouring and proposes a liquid transfer algorithm based on a parametric hidden Markov model.     
The difficulty of generating pouring motion increases when pouring accuracy is essential. The demand for accurate pouring is observed in casting factories where molten metal is poured into molds. \cite{7068564} proposes predicting the residual pouring amount of the liquid to increase the accuracy of pouring. \cite{4758180} introduces predictive sequence control, which suppresses the increase of error when the pouring amount increases. 

Humans also control the amount when they pour and for which they combine pouring with shaking and tapping \cite{doi:10.1142/S0219843615500309}. Efforts have been made to estimate the volume or height of the poured amount in the receiving container and to use the estimate as real-time feedback to a simple PID controller. \cite{7989307} uses a deep neural network to estimate the volume of liquid in a cup using visual data and uses PID controllers to control the rotation of a robot arm. In 30 pours, the algorithm achieves an average error of 38mL with 25 seconds per pour. \cite{do2018accurate} uses an RGB-D point cloud of the receiving cup to determine the liquid height and a PID controller to control the rotating angle of the source cup. They achieve a mean error of 23.9mL, 13.2mL, and 30.5mL for three different receiving cups. In both algorithms, the PID controller stops and rotates the source container back to the original angle when the estimated volume/height reaches the target. However, the before-mentioned technique might lead to over-pouring since there is still liquid coming out from the source container when it starts its backward rotation as shown in Figure \ref{fig-human-pouring}. Moreover, the generalization of those vision-based approaches is limited when there is variation in the color of the receiving container, the lighting conditions, the background, or the type of pouring material.

Instead of using PID, \cite{chaudo2018} learns a policy using reinforcement learning simulation and transfers the policy to actual robots. The policy performs pouring to the same target heights for which it was trained.  It reaches an average error of 19.96mL over 40 pouring trials. However, the authors also use a vision-based system to detect the height of the liquid in the receiving container, leading to the same limitations already discussed. 
In \cite{hong2019pouring_audio} the authors rely on an audio spectrogram to determine the volume poured by the robot. The mean volume errors reported for different receiving containers ranged from 6.42\,ml to 13.79\,ml, such small errors are achieved by the usage of a spout at the opening of the pouring containers, which reduces the speed of pouring.
\cite{8202301, 8653969} derives analytical pouring models for the source containers with known geometry and extends the model to source containers with similar geometry. The most up-to-date work \cite{8653969} uses both vision and weight during pouring and achieves a pouring error of less than 5mL. However, the proposed system pouring time ranges from 20 to 45 seconds. Humans can also achieve small pouring errors if requested to pour slowly. In our dataset, humans took 3.2 to 8.7 seconds to pour water. The authors in \cite{tianze2019} apply model predictive control (MPC) based on a recurrent neural network for estimating the poured volume. The approach achieves average errors of 14.25mL, 18.25mL, and 26.13mL for three unseen source containers. Another work \cite{huang2017learning_pour} presented a Long-Short-Term Memory (LSTM) model that was trained using demonstration data. However, the learned model was only evaluated in simulation. 

In summary, the accurate pouring approach presented in this paper has a significant improvement over our previous works. The proposed peephole LSTM approach drastically outperforms the MPC algorithm presented in \cite{tianze2019}. The proposed approach was applied to a real robot, whereas in \cite{huang2017learning_pour} the pouring motion velocities were generated in simulation. In  \cite{huang2019accurate}, we presented limited preliminary results on pouring as an abstract report. A generalization in practice (GiP) approach was proposed in \cite{wilches2020generalizing} with limited experiments on pouring water. This paper presents a comprehensive description of the peephole LSTM pouring motion generation approach and the generalization by self-supervised practicing (GSSP), both thoroughly evaluated with numerous experiments in real-world scenarios with a real robotic system.  

\section{Problem Description \& Approach}

In this work, we do not consider the transfer of the source container, which is essentially pick-and-place, and only focus on controlling the flow of the liquid by manipulating the source container. We consider the receiving container to be large enough to prevent spilling. 
The main movement of the source container is its rotation, which resides mostly on a 2-dimensional plane.  It enables the motion to be simplified to its rotation. The anchor of the rotation is fixed to be approximately at the middle point of the height of each container. 
This simplification is also applied in \cite{7989307, 8653969, do2018accurate, 8202301}, which makes our assumptions reasonable. 
Volume can be perceived visually and is intuitive for measuring liquid. Therefore, in this work, we use volume to represent the amount of liquid.  

We describe the pouring process as a result of rotating the source container. Initially, a certain volume of liquid exists in the source container. If the source container is full, then the liquid flows out as the source container starts rotating. 
If the source container is not full, then there will be a delay of the liquid flowing out after the source container has started rotating.
The liquid flows into and stays in the receiving container, and therefore the poured volume only increases and never decreases. When the source container stops rotating, the liquid may either instantly stop flowing out or keep flowing for a short time until the surface of the liquid inside the source container is level. The pouring process is sequential, and the poured volume is determined by the trajectory of the rotation velocities of the source container.

\begin{figure}
\centering
\includegraphics[width=0.95\columnwidth]{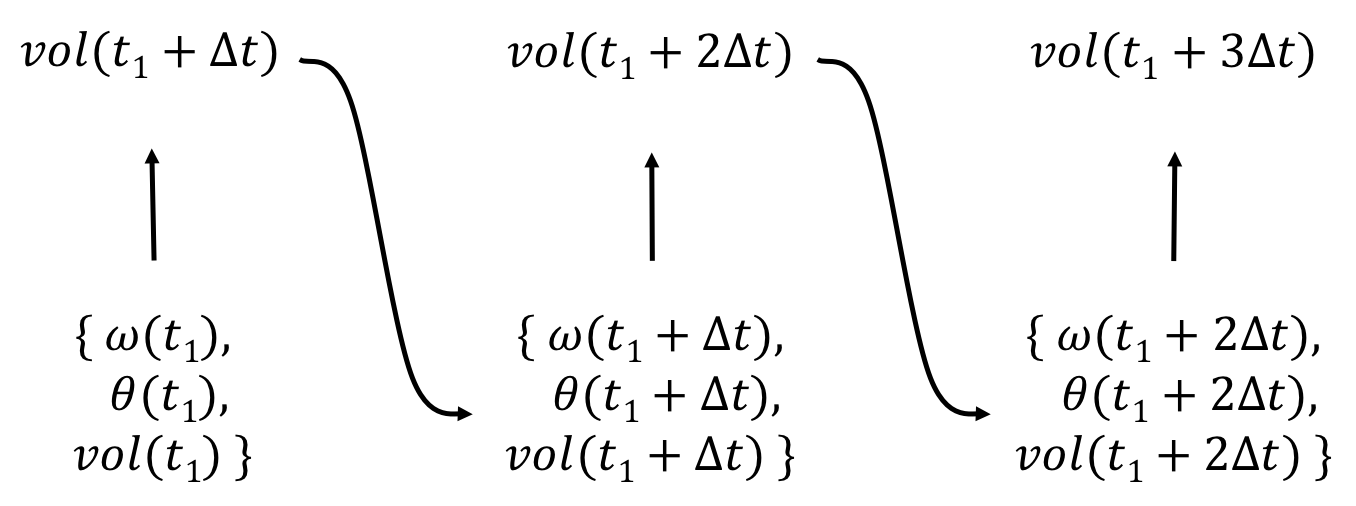}
\caption{The sequential pouring process with the input being the current angular velocity and the current poured volume and the output being the poured volume for the next time step}
\label{fig-system_illus}
\end{figure}

We model the pouring process  as a discrete-time series:
\begin{algorithmic}[1]
\For {$i$ in ($1, 2, \dots$) }
	\State $t = t_1 + (i-1)\Delta t$
	\State $\theta(t + \Delta t) = \theta(t) + \omega(t)\Delta t$
	\State $vol(t + \Delta t) = F(\omega(t),\theta(t),vol(t))$
\EndFor    
\end{algorithmic}  
where $t_1$ is the initial time instant, $\Delta t$ is the time interval, $\theta(t)$ and $\omega(t)$ are the rotation angle and angular velocity of the source container, respectively, $vol(t)$ is the poured volume,  $F(\cdot)$ denotes the pouring system. We also illustrate the process in Figure \ref{fig-system_illus}. We do not impose a strict restriction on the initial angle $\theta(t_1)$ but assume that it is close to zero.
The effect of the velocity $\omega(t)$ executed at time $t$ is observed at the next time step, $t + \Delta t$, and the effects are the next rotation angle $\theta(t + \Delta t)$ and the next poured volume $vol(t + \Delta t)$. Other factors that affect the pouring behavior considered in this paper are the shape of the source container, the initial volume in the source container, and the target volume in the receiving container.

The angular velocity, $\omega(t)$, is the action that pushes the pouring process forward. To perform pouring, we can use a motion model that takes the target volume as input and generates the velocity as output. At any time step during pouring, the model takes the current poured volume as input, compares it with the target volume, and adjusts the velocity accordingly. 
The model is represented as 
\begin{equation} \label{eq-vel_gen}
\omega(t) = G(\omega(t - \Delta t), \theta(t), vol(t), vol_{2pour}),
\end{equation}
where $G(\cdot)$ denotes the function that relates the previous velocity $\omega(t-\Delta t)$, current angle $\theta(t)$ and volume $vol(t)$, and target volume $vol_{2pour}$ with the current velocity. The pouring process is written again as:
\begin{algorithmic}[1]
\For {$i$ in ($1, 2, \dots$) }
	\State $t = t_1 + (i-1)\Delta t$
	\State $\omega(t) = G(\omega(t - \Delta t), \theta(t), vol(t), vol_{2pour})$ 
	\State $\theta(t + \Delta t) = \theta(t) + \omega(t)\Delta t$
	\State $vol(t + \Delta t) = F(\omega(t),\theta(t),vol(t))$
\EndFor    
\end{algorithmic}  

\subsection{Self-Supervised Learning from Demonstration and Outcomes}

The learned pouring model should have the targeted volume as one of the inputs since the pouring behavior should change based on the target.  However, as one of the critical innovations in our approach, we replace the targets with the actual results in training. For example, we ask subjects to pour 150mL, but the subject pours 165mL.  In this case, the input to the model is 165mL, not 150mL.  This change enables the proposed self-supervised learning since the subjects in the demonstration do not have to be supervised, and the desired target does not have to be labeled. The robot can observe the actual outcomes and use them as training inputs. This approach allows the model to pour with a human-like pace but achieves better accuracy. 

In the model, although taking $vol_{2pour}$ as input, $G(\cdot)$ is not guaranteed to generate the exact $\omega(t)$'s which will lead to $vol_{2pour}$. In reality, given $vol_{2pour}$, $G(\cdot)$ will generate $\omega(t)$'s whose execution will lead to a certain final volume $vol_{final}$ in the receiving container. Assume the receiving container initially has a volume $vol_{init}$, then $vol_d =| vol_{final} - vol_{init} - vol_{2pour}|$ reflects the ability of $G(\cdot)$ to fulfill $vol_{2pour}$. If $vol_d = 0$, then $G(\cdot)$ is perfect in fulfilling the goal. Reversely, $vol_{final}$ can be considered as the result of executing a \emph{perfect} model $G^*(\cdot)$ assuming its goal is set to $vol_{2pour} = vol_{final} - vol_{init}$. 
To learn $G(\cdot)$, if we use the actual $vol_{2pour}$, i.e., the volume we \emph{intend} to reach, then the learned model will approximate the one that is given by $vol_{2pour}$ but ends with $vol_{final}$. If we set $vol_{2pour} = vol_{final} - vol_{init}$, the learned model will approximate the perfect model $G^*(\cdot)$. In the hope of learning a more accurate motion model, we set the motion goal $vol_{2pour}$ using the actual outcome $vol_{final} - vol_{init}$. 

\subsection{RNN-Based Pouring Skill Model}

A straightforward design choice for accurate pouring is a simple PID controller. \cite{7989307, do2018accurate} have applied PID for accurate pouring, in which they both used PIDs with fixed gains. While pouring, as the liquid flows out of the source container, the plant changes, which accordingly requires adjustment to the gains of the PID. Performing pouring thus requires an adaptive PID whose gains change their values throughout the pouring process. We speculate that this partly limits the achievable accuracy in \cite{7989307, do2018accurate}. An adaptive PID, however, is no longer a simple controller. That justifies our quest for a more complicated motion model.

We aim to learn a motion model from pouring demonstrations that can perform properly in new settings after being trained in a finite number of settings. In this work, we explore the generalization of the pouring skill model to different shapes of source containers, type of liquid, and granular materials. Generalization of neural networks has been observed in practice, and active research has been conducted, which tries to identify possible causes such as the norm of network parameters \cite{NIPS2017_7176}, the specialty of the network structure and the landscape of the cost function \cite{DBLP:journals/corr/WuZE17}, and sharpness/flatness of the minima \cite{Dinh:2017:SMG:3305381.3305487}. 

Apart from generalization, we seek two other properties from the candidate model:
\begin{enumerate}
\item Since all demonstrations are sequences, the model should be inherently capable of dealing with sequences and capturing the spatial-temporal patterns in the sequences.
\item Since demonstrations vary in length, the model should be able to learn effectively from sequences with different lengths. 
\end{enumerate} 

Due to the successful records of the generalizability of neural networks and our need for a sequential model, we use RNN to represent the motion model. 
RNN is a class of neural networks that is designed to process its inputs in order. It feeds its output from one time step into its input at the next time step, shown specifically in Eq. \eqref{eq-rnn}, where $x(t)$ is the given input, $h(t-1)$ and $h(t)$ are outputs from the previous and the current step, respectively. The weight $W$ and bias $b$ are learned using Backpropagation Through Time \cite{werbos1990}.  
\begin{equation} \label{eq-rnn}
h(t) = \text{tanh}\left(W[h(t-1), x(t)]^\top + b\right)
\end{equation}

\begin{figure}
\centering
  \includegraphics[width=0.95\columnwidth]{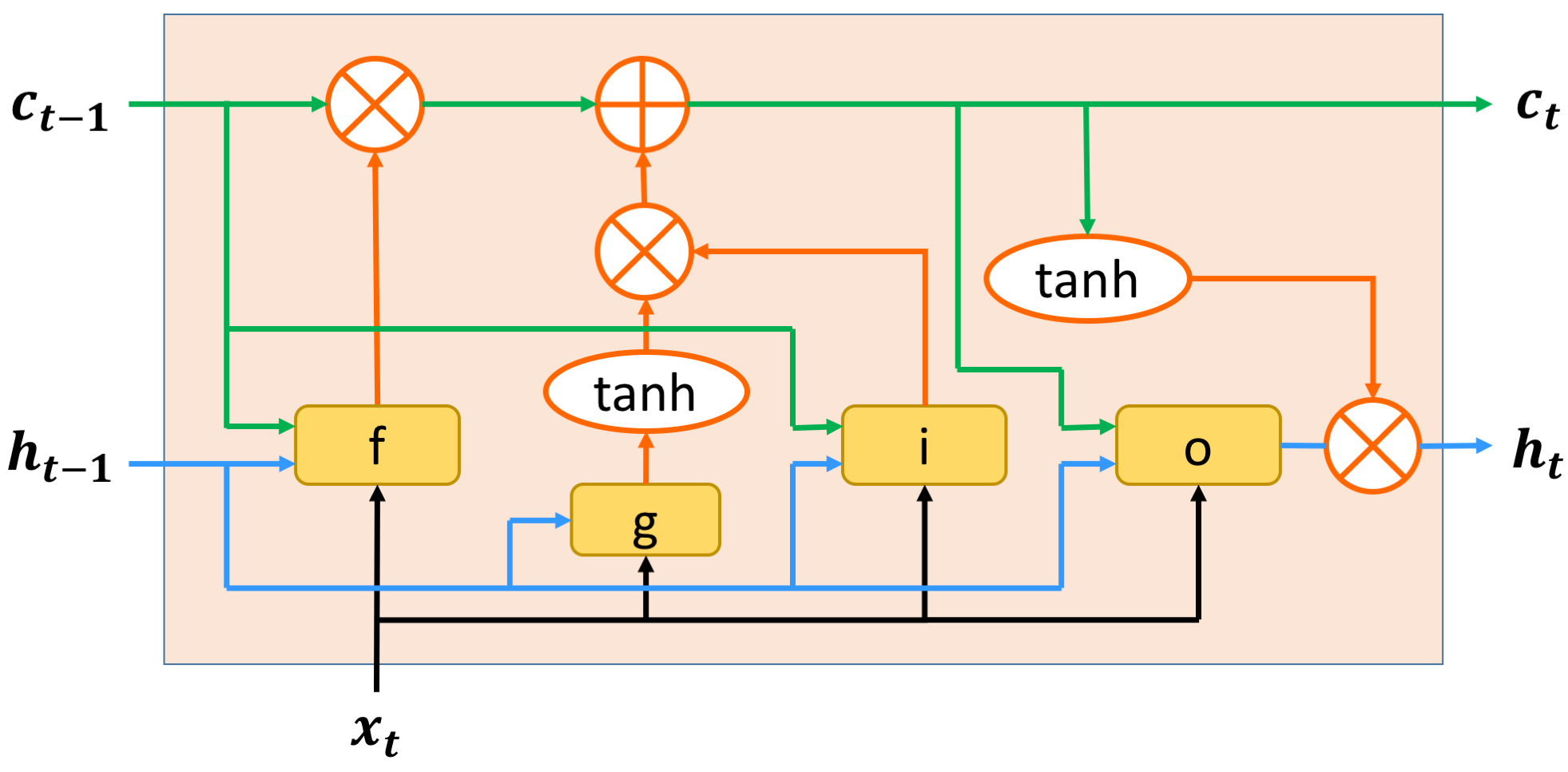}
    \caption{Mechanism of a peephole LSTM block}
    \label{fig-lstm}
\end{figure}

\begin{figure}
\centering
\includegraphics[width=0.95\columnwidth]{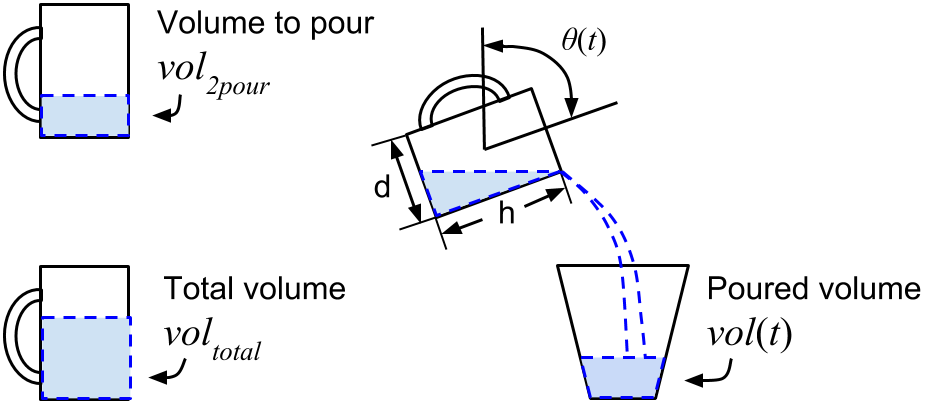}
\caption{An illustrative pouring scene that shows the six physical quantities to obtain. $vol_{2pour}$ and $vol_{total}$ is the target and initial volume. $D$ and $H$ are the diameter and height of the source container. $\theta(t)$ and $vol(t)$ are the sequences of rotation angle and of the poured volume.}
\label{fig-pouring_scene}
\end{figure}

We need to decide the input features to the RNN at any time step. Each feature corresponds to a type of data. We write Eq. \eqref{eq-vel_gen} again below for convenience:
\begin{equation}
\omega(t) = G(\omega(t - \Delta t), \theta(t), vol(t), vol_{2pour})
\end{equation}
The first feature is the previous angular velocity that can be encoded as the hidden state of the RNN represented by Eq. \eqref{eq-rnn}. We also use $\theta(t)$ as a feature. 
The next two features are $vol(t)$ and $vol_{2pour}$, respectively. 
Corresponding to $vol_{2pour}$, the initial volume of liquid in the source container $vol_{total}$ can be set as a feature. We can also have features that describe the shape of the source container. We model the source container as a cylinder and set both the height $H$ and the body diameter $D$ as features.
The four static features $vol_{2pour}$, $vol_{total}$, $H$, and $D$ describe a pouring task and distinguish one task from another. The two sequential features $\theta(t)$ and $vol(t)$ represent the feedback from the rotation on the source container and the volume change on the receiving container. Figure \ref{fig-pouring_scene} illustrates the six input features. Therefore, the input and output of the RNN from Eq. \eqref{eq-rnn} become:
\begin{equation} \label{input-rnn}
x(t)=[\theta(t), vol(t), vol_{total}, vol_{2pour}, H, D]
\end{equation}
\begin{equation} \label{output-rnn}
\omega(t) = K(h(t))
\end{equation}
where the function $K(.)$ relates the hidden state of the RNN to the scalar angular velocity. The plain RNN as shown in Eq. \eqref{eq-rnn} suffers from the problem of vanishing and exploding gradients \cite{bengio1994, hochreiter1997}, which prevents it from learning long-term dependencies effectively. The problem was solved by long short-term memory (LSTM) which introduces gates and memory cells \cite{hochreiter1997}. Later, peepholes were introduced to LSTM to enable the access of all gates to the memory cell \cite{Gers:2003:LPT:944919.944925}. The mechanism of peephole LSTM is illustrated in Figure \ref{fig-lstm} and is written as:

\begin{align}
	i &= \text{sigm}\left(W_i[h(t-1), x(t)]^\top + b_i + p_i\odot c(t-1) \right) \label{lstm-input} \\
    f &= \text{sigm}\left(W_f[h(t-1), x(t)]^\top + b_f + p_f\odot c(t-1) \right) \label{lstm-forget} \\
    g &= \text{tanh}\left(W_g[h(t-1), x(t)]^\top + b_g \right) \label{lstm-input1}\\
    c(t) &= f \odot c(t-1) + i \odot g \label{lstm-cell}\\
    o &= \text{sigm}\left(W_o[h(t-1), x(t)]^\top + b_o + p_o\odot c(t) \right) \label{lstm-output}\\
    h(t) &= o \odot \text{tanh}(c(t)) \label{lstm-hidden}
\end{align}      
where $i$, $o$, and $f$ are the input, output, and forget gates, respectively. $W_i$, $W_f$, $W_g$, $W_o$, $b_i$, $b_f$, and $b_g$ are the LSTM cell weights to learn. $p_i$, $p_o$, and $p_f$ are also the peephole connection weights to be learned for gates $i$, $o,$ and $f$, respectively. $c(t)$ is the long-term memory, $h(t)$ the output, and $x(t)$ the input of the LSTM block. ``sigm" represents the sigmoid function applied element-wise, and is used to implement gates. ``tanh" represents the hyperbolic tangent function, applied element-wise, and is used to avoid vanishing or exploding gradients. $\odot$ represent element-wise multiplication. In this work, we use peephole LSTMs.

Taking into account \cref{input-rnn,output-rnn,lstm-input,lstm-forget,lstm-input1,lstm-cell,lstm-output,lstm-hidden} we can see that the dynamic model of the pouring motion is allocated in the long-term memory $c(t)$ of the LSTM. The combination of Eq. \eqref{lstm-cell} and Eq. \eqref{lstm-input1} gives:
\begin{equation}
c(t) = f \odot c(t-1) + i \odot \text{tanh}\left(W_g[h(t-1), x(t)]^\top + b_g \right)
\end{equation}
where the dynamic model $c(t)$ depends on both the previous dynamics and the previous input and output. The gate $i$ decides which part of the current input and past output contributes to the current dynamic model, and the gate $f$ decides which part of the past dynamic model contributes to the current dynamic model. Eq. \eqref{lstm-hidden} shows that gate $o$ decides which part of the dynamic model will be used as the current output. The previous mechanism allows the LSTM network to predict when to start the backward rotation of the source container based on the information provided by its current input (feedback signals and static features), its past output (past angular velocity) and its long-term memory (past dynamic model).

\section{Training} \label{sec:training}

\subsection{Data Collection} \label{sec:dataset}

\begin{figure}
\centering
\includegraphics[width=0.95\columnwidth]{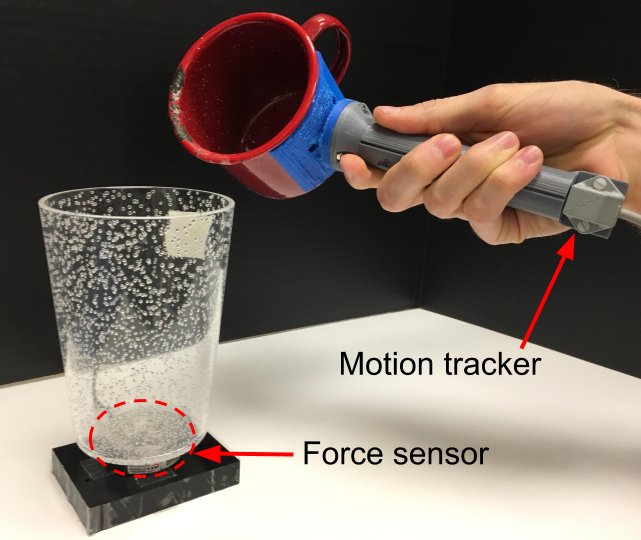}
\caption{Illustration of the data collection setup. The source container is connected to the motion tracker through a 3-D printed adapter. The force sensor is placed underneath the receiving container.}
\label{fig-data_collect_setup}
\end{figure}

\begin{figure}
\centering
\includegraphics[width=0.95\columnwidth]{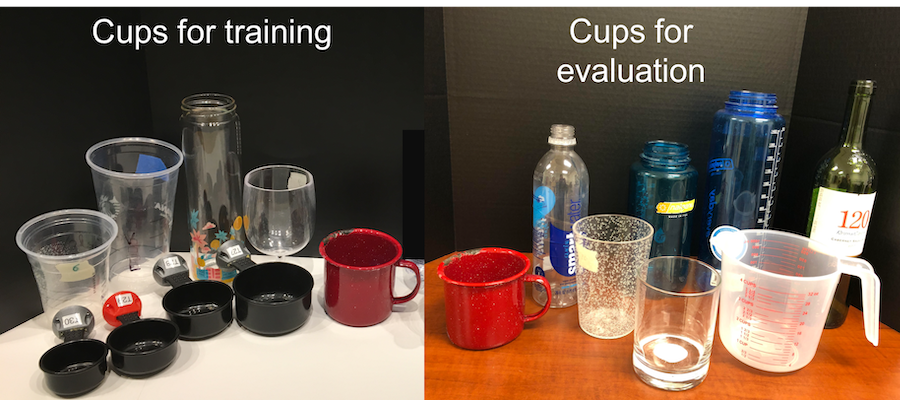}
\caption{Source containers used for (left) training and for (right) evaluation. The red cup was used both for training and for evaluation.}
\label{fig-cups}
\end{figure}

We want to collect all the input features that we have identified for RNN and we need to decide how to measure volume. Intuitively, the volumes $vol_{total}$ and $vol_{2pour}$ can be measured using a measuring cup. However, obtaining $vol(t)$ using a measuring cup requires a real-time video stream of the measuring cup and a computer vision algorithm that extracts the volume from the video stream.   
To simplify the problem that we have to solve, we decide that we will not include the above vision problem in our solution, and instead, we compute the volume from other quantities. 
The volume can be computed as the mass $m$ divided by the density $\rho$, i.e., $v = m / \rho$. We consider the weight as the gravitational force acting on an object that keeps the object in place. The weight $f$ is the product of mass $m$ and gravitational acceleration $g$, i.e., $f = mg$. Therefore, the volume can be calculated from the weight:  
\begin{equation} \label{eq-volume_weight}
v = \frac{f}{\rho g},
\end{equation}
We represent $vol_{total}$ by its corresponding weight $f_{total}$, $vol_{2pour}$ by weight $f_{2pour}$, and similarly the current poured volume $vol(t)$ by weight $f(t)$. 
Figure \ref{fig-data_collect_setup} illustrates the setup for our data collection. We collect data of pouring water from 9 different source containers into the same receiving container. The 9 source containers are shown as the left half of Figure \ref{fig-cups}.  We measure $H$ and $D$ of each source container in millimeters (mm) using a ruler. We 3D-print a handle where the source container is mounted on one end, and a Polhemus Patriot motion tracker is mounted on the other end. The motion tracker records the rotating angles $\theta(t)$'s of the source container in degrees. We place an ATI Mini40 force/torque sensor under the receiving container to record the raw force reading $f_{raw}(t)$ in pound-force (lbf).  

We obtain $f_{total}$ and $f_{2pour}$ from $f_{raw}(t)$. $f_{2pour}$ is calculated by $f_{2pour} = f_{final} - f_{init}$, where $f_{init}$ and $f_{final}$ are the weights read from the receiving container before and after a trial, respectively. Thus, we set $f_{2pour}$ using the actual poured outcome.
In each trial, $f_{total} > f_{2pour}$, that is, there is water left in the source container after pouring. Various $f_{total}$ and $f_{2pour}$ are recorded to aid the generalizability of the prospective motion model.  
$\theta(t)$'s are recorded at 60Hz and $f_{raw}(t)$'t are recorded at 1KHz. The collected pouring data is part of RPAL Daily Interactive Manipulation (DIM) dataset \cite{huang2019dataset}, which is publicly available.  More manipulation datasets can be found at \cite{huang2016recent}. 

\subsection{Implementation} \label{sec:implementation}

The neural network can have multiple layers and each layer can contain multiple peephole LSTM units. By units we mean the size of the vectors $\mathbf{c}_t$ and $\mathbf{h}_t$ of Figure \ref{fig-lstm} formalized by Eqs. \eqref{lstm-cell} and \eqref{lstm-hidden}, respectively. Dropout \cite{zaremba2014} is applied between layers to avoid memorizing the data and aid generalizability. 
The final layer is a fully connected layer with linear activation which generates the angular velocity. The mechanism of the network with $L$ layers at time $t$ is represented as
\begin{algorithmic}[1]
\State $h_0(t) = x(t)$
\For {$i=(1, 2, \dots, L)$} 
	\State $h_i(t) = \text{LSTM}\left(h_i(t-1), h_{i-1}(t); n_{unit}\right)$
	\State $h_i(t) = \text{Dropout}\left(h_i(t); p_{keep}\right)$ 
\EndFor
\State $\hat{y}(t) = W_yh_L(t) + b_y$
\end{algorithmic}
where $\text{LSTM}(\cdot; n_{unit})$ means using an LSTM block from Figure \ref{fig-lstm} with $n_{unit}$ units. Dropout$(\cdot; p_{keep})$ means dropout with a keep probability of $p_{keep}$. $\hat{y}(t)$ corresponds to a linear layer that converts the hidden state $h_L(t)$ to the final output.

To feed the input features into the network, we group them into a vector $x(t)=[\theta(t), f(t), f_{total}, f_{2pour}, H, \kappa]^\top$ for $t=1,\dots,T-1$, where $T$ is the length of the trial and  
\begin{enumerate}
\item $\theta(t)$ is the rotating angle of the source container.
\item $f(t)$ is the weight of the poured liquid.
\item $f_{total}$ is the weight of the initial amount of liquid present in the source container before pouring. 
\item $f_{2pour}$ is the weight of the target poured amount. 
\item $H$ is the height of the source container.
\item $\kappa$ is the body curvature of the source container.  
\end{enumerate}
The body curvature $\kappa$ of the source container is calculated from the body diameter, $D$:
\begin{equation}
\kappa = 2 / D
\end{equation}
The angular velocities $\omega(1:T-1)$ are computed from $\theta(1:T)$:
\begin{equation}
\omega(t) = (\theta(t+1) - \theta(t))f_s, \qquad t=1, 2, \dots, T-1
\end{equation}    
where $f_s$ is the sampling frequency of $\theta(t)$. 
For each trial, at time $t\in[1, 2, \dots, T-1]$, the input $x(t)$ and target $y(t)$ of the network are
\begin{align}
x(t) &= [\theta(t), f(t), f_{total}, f_{2pour}, H, \kappa]^\top\\
y(t) &= \omega(t)
\end{align}     
The output of the network is denoted by $\hat{y}(t)$. Assume we have $N$ trials in total, and each trial has length $T_i$, $i\in[1, 2, \dots, N]$. The loss function is defined as 
\begin{equation}
c = \frac{1}{N}\sum_{i=1}^N\frac{1}{T_i-1}\sum_{t=1}^{T_i-1}(\hat{y}_i(t) - y_i(t))^2.
\end{equation}

\subsection{Data Preparation} 
 
\begin{figure}
\centering
\includegraphics[width=0.95\columnwidth]{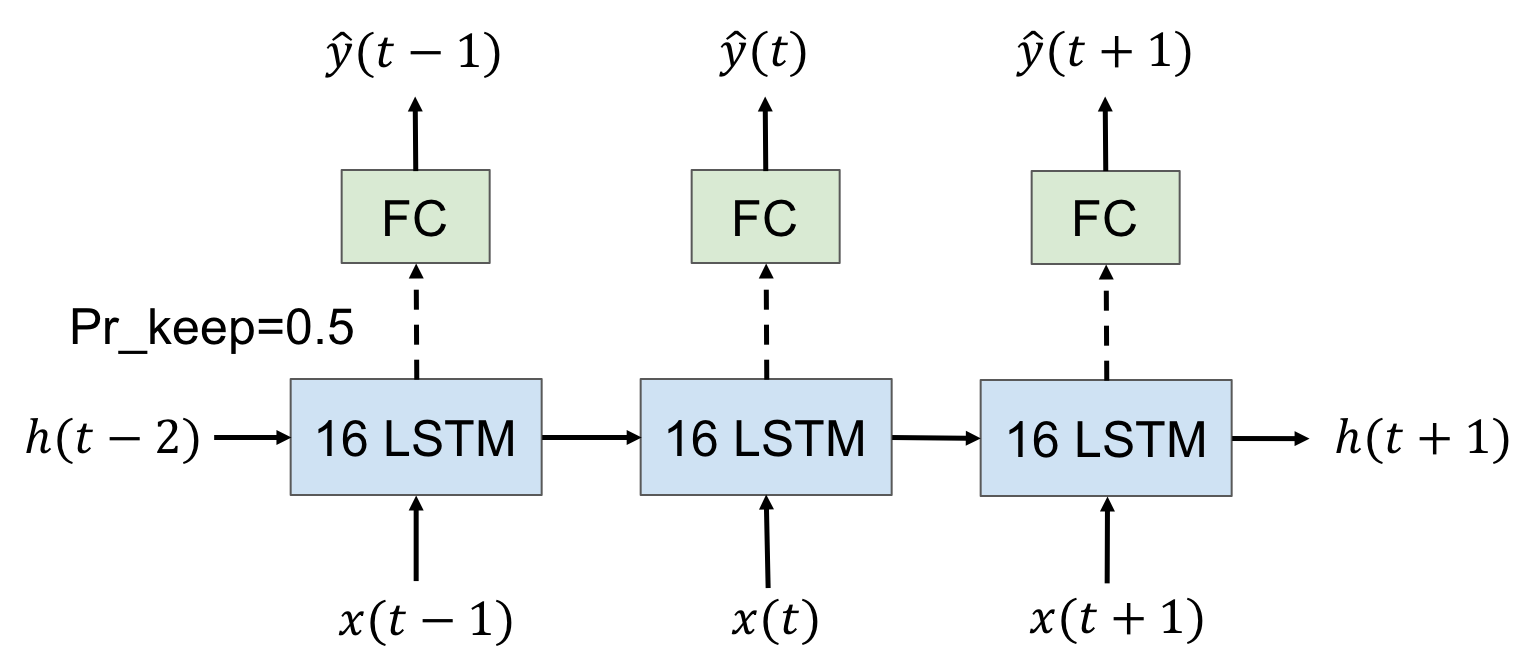}
\caption{Our network has 1 layer and 16 peephole LSTM units. Dropout with a keep probability of 0.5 is applied to non-sequential connections. }
\label{fig-network}
\end{figure}

We set the sampling frequency $f_s=60$Hz since it is the lower one between the frequencies of $\theta(t)$ and $f_{raw}(t)$. We kept the recorded $\theta(t)$'s intact and downsampled $f_{raw}(t)$ to 60Hz. We obtain $f(t)$ by filtering the raw reading from the force sensor $f_{raw}(t)$, specifically
\begin{align}
f_m(1:t) &\leftarrow \text{median}\_\text{filter}(f_{raw}(1:t)), \quad \text{window}\_\text{size}=5, \\
f(t) &\leftarrow \text{Gaussian}\_\text{filter}(f_m(1:t)), \quad \sigma=2.
\end{align} 
We normalize each input dimension independently using the mean and standard deviation of that dimension. 
The model had 1 layer and 16 LSTM units. We trained models with different numbers of layers and LSTM units, and we found the model with 1 layer and 16 units had a simple structure and performed well. We set the keep probability of dropout to be 0.5. Specifically, the computation for time step $t$ is represented as:
\begin{align}
h(t) &= \text{LSTM}\left(h(t-1), x(t)\right) \\
h_d(t) &= \text{Dropout}\left(h(t)\right) \\ 
\hat{y}(t) &= W_yh_d(t) + b_y
\end{align}
The network is shown in Figure \ref{fig-network}. 

The learning model involves 284 trials in total, among which 221 are for training and 63 for validation. Each iteration is an epoch, in which the entire training and validation data are traversed. We ran 2000 epochs and picked the model that has the lowest validation loss. We used the Adam optimizer and set the initial learning rate to be 0.001. The code is written using TensorFlow. 

\section{Generalization by Self-Supervised Practicing}

The robot with the motion model has limited generalizability related to certain pouring containers for which it achieved high pouring error. We recorded the outcomes that the robot provided when pouring with the poor-performing containers. We used those outcomes to fine-tune the model. We refer to this fine-tuning process using the outcomes generated by the robot as generalization by self-supervised practicing (GSSP).
GSSP updates the motion model by training it with data generated by practices. Although the outcomes in the first few practices may present a high volume error, they contain the response of the model to new conditions and are therefore valuable. GSSP mimics the behavior of humans when they face a new instance of the pouring task.  When we use a cup that we are not familiar with, we pour based on our experience. Shortly after a few practices, the model improves and adapts to the new cup. The key component in GSSP is that the robot uses the actual outcomes instead of the desired outcomes to fine-tune the model. 

We refer to the process of using the model on a robot to finish a pouring trial as a {\it practice}. We use the robot's outcome sequences that result from several practices to form a new dataset, which we use to fine-tune the model. The fine-tuning also uses the actual outcomes as the motion goal. The inputs for the fine-tuning process are: 
\begin{itemize}
\item $H$: height of the source container.
\item $\kappa = 2 / D$: body curvature of the source container with $d$ being the diameter of the container.
\item $f_{total}$: the initial weight of water present in the source container before pouring. 
\item $f_{2pour}$: the actual weight of water that was poured during the practice. 
\item $f(t)$: the sequence of the weight of water in the receiving container in the practice.
\item $\theta(t)$: source container's current angle during the practice.
\end{itemize}
The output of the fine-tuning is:
\begin{itemize}
\item $\omega(t)$: angular velocity of the source container \textbf{in the practice}.
\end{itemize} 
 
Before fine-tuning, we obtain an initial model through the presented self-supervised learning from demonstration. The fine-tuning process in GSSP can be carried out in two ways:
\begin{itemize}
    \item \textbf{Gradual Fine-tuning}: First, the robot performs $n$ practices, where $n$ is relatively small. The resulting pouring sequences form a new dataset, which is used to fine-tune the model. If the error of the updated model is larger than a predefined threshold, $n$ practices are performed again, the newly generated $n$ data points are added to the new dataset. The practices keep being performed and the new dataset keeps growing until the error of the updated model is below the predefined threshold.
    We define the error threshold to be two times that of an averaged human's pouring error. Algorithm \ref{alg:gip} describes gradual fine-tuning with $n \in [5,15]$.
    \item \textbf{Batch Fine-tuning}: First, the robot performs $n$ practices, where $n$ is relatively large. The resulting pouring sequences are used to fine-tune the model. Batch fine-tuning is equivalent to conducting one iteration of gradual fine-tuning with a large $n$, i.e., $n > 35$, where 35 is the average number of samples per source container for the initial training set discussed in section \ref{sec:training}. 
\end{itemize}

\begin{algorithm*}[h!]
\caption{Generalization by self-supervised practicing (gradual fine-tuning)}\label{alg:gip}
\begin{algorithmic}[1]
\State $M_{init}$: Initial model
\State $n$: Number of practices
\State $\mathcal{R}\gets$ \{($r^1_s$, $r^1_g)$, ..., $(r^n_s$, $r^n_g$)\}
 \Comment{$r^i_s$: start state; $r^i_g$ desired outcome or goal}
\State $err\_th\gets$ error threshold
\State $\mathcal{D}\gets \{\}$

\Procedure{Practice}{$M$, $n$, $\mathcal{R}$}
\Repeat
    \State Robot practices once using one item in  $\mathcal{R}$ using model $M$
    \State $\mathcal{D}\gets \mathcal{D} \cup \{d\}$ \Comment$d:$ outcome sequence from the practice 
\Until {$n$ practices have been performed}
\State error $\gets$ mean error between actual outcome and desired outcome among $n$ practices
\State \textbf{return} error \label{return}
\EndProcedure

\Procedure{GSSP}{}
\State $M_{new} \gets M_{init}$
\While {True}
    \State $err$ $\gets$ \Call {Practice}{$M_{new}$, $n$, $\mathcal{R}$}
    \If {$err < err\_th$}
        \State \textit{break}
    \Else
        \State $M_{new} \gets$ {Fine-tune}($M_{new}, \mathcal{D}$)
        \Comment{Fine-tune the model using $\mathcal{D}$}
        \State $\mathcal{R} \gets$ {Generate-Random-Practices} ($n$) 
        \Comment{Randomly generate $n$ set of requirements for future practices}
    \EndIf
\EndWhile
\State \textbf{return} $M_{new}$
\EndProcedure

\end{algorithmic}
\end{algorithm*}

The GSSP approach can be used for any manipulation where the actual outcomes of several practices can be used as the desired outcomes, i.e., the actual outcome can substitute the desired outcome for training. Taking throwing objects into bins as an example  \cite{zeng2020tossingbot}, we can practice throwing with the robot using unseen objects, record the outcome, and apply GSSP to generalize the learning model to new objects. Another example can be recording the state of food ingredients \cite{jelodar2018identifying}, or a state change \cite{jelodar2019joint} after the execution of a manipulation, record the action sequences, use them as training samples, and fine-tune the manipulation model to expand its generalization. However, the GSSP approach may not generalize well among different manipulations. For example, a throwing motion model may not be generalized to mixing manipulations since the two motion models may have different model structures.  Our latest work on motion code and motion embedding \cite{paulius2020motion, alibayev2020estimating} may help in this kind of cross-motion-type generalization. 

\section{Experiments \& Evaluation} \label{sec:experiments}

\begin{figure}[h]
\begin{center}
\includegraphics[width=0.61\linewidth]{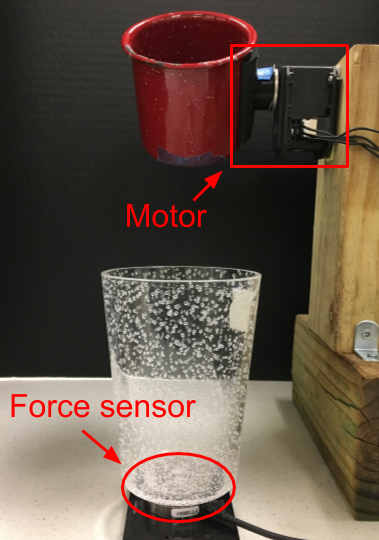}
\includegraphics[width=0.37\linewidth]{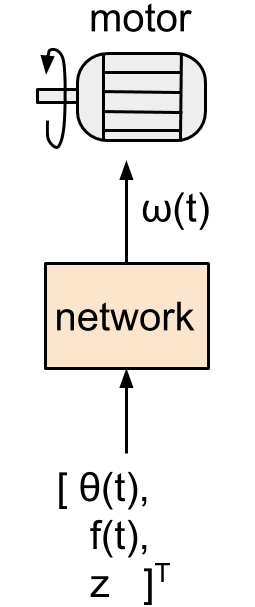}
\end{center}
\caption{(Left) The robotic system consists of a motor that executes the generated velocity command and a force sensor that monitors the poured amount. The source containers are attached to the motor through a 3-D printed adapter. (Right) Before pouring, we obtain the static features $z=[f_{total}, f_{2pour}, H, \kappa]$. At time step $t$, the robotic system obtains $\theta(t)$ and $f(t)$, combine them with $z$, and send to the network. The network generates velocity command $\omega(t)$ which is executed by the motor. 
}
\label{fig-physical_system}
\end{figure}

To evaluate the motion model, we built a robotic system that consists of the trained RNN, a Dynamixel MX-64 motor, and the same force sensor with which we collected the data. The motor was placed at a certain height above the surface. The force sensor was placed on the surface close by. The source container was attached to the motor. The receiving container was placed on top of the force sensor. We properly placed the receiving container (along with the force sensor) according to the particular source container used so that there is little spilling. Figure \ref{fig-physical_system} (Left) shows the setup of the robotic system. 
It runs at 60Hz, the same as the data collection. The time between consecutive time steps is $\Delta t = 0.016$ seconds. Before performing each separate pouring trial, we obtain the four static features which we denote by $z=[f_{total}, f_{2pour}, H, \kappa]$. During the trial, at time step $t$, we obtain $\theta(t)$ from the motor and $f(t)$ from the force sensor, and we feed the input features $x(t) = [\theta(t), f(t), z]^\top$ to the model, which then generates the velocity $\omega(t)$. The motor executes the velocity. The above process repeats at time step $t + \Delta t$. Figure \ref{fig-physical_system} (Right) shows the working process of the robotic system at time $t$.  

The robotic system:
\begin{enumerate}
\item Normalized every input dimension. 
\item Obtained $f(t)$ by filtering the raw force readings. 
\end{enumerate}
in the same way as in training.
We evaluated the motion model by testing it on pouring certain kinds of liquid from certain source containers. The difficulty of the task changes when the liquid and the source container change. For each pair of liquid and source containers, the model pours 15 times, each time with arbitrary $vol_{total}$ and $vol_{2pour}$, where $vol_{total}>vol_{2pour}$. We show the pouring error of each pair of liquid and source container in the form of figures. In the figure, we plot the actual poured volume against the target volume for all 15 trials. We also show the liquid type,  the mean,  and standard deviation of the pouring error: $\mu_e$ and $\sigma_e$ in milliliters. At the bottom right of the figure, we show the source container that was used. We also show a black dashed line that illustrates zero pouring error.
Translating the force reading to volume requires the density of the liquid $\rho$ and the gravitational acceleration $g$. We used 0.997g/mL for the density of water and 9.80665 m/s$^2$ for gravitational acceleration.

\subsection{Model Evaluation} \label{sect:eval1}

\subsubsection{Model Evaluation of Pouring Water}

\begin{figure}[h]
\begin{center}
\includegraphics[width=0.95\columnwidth]{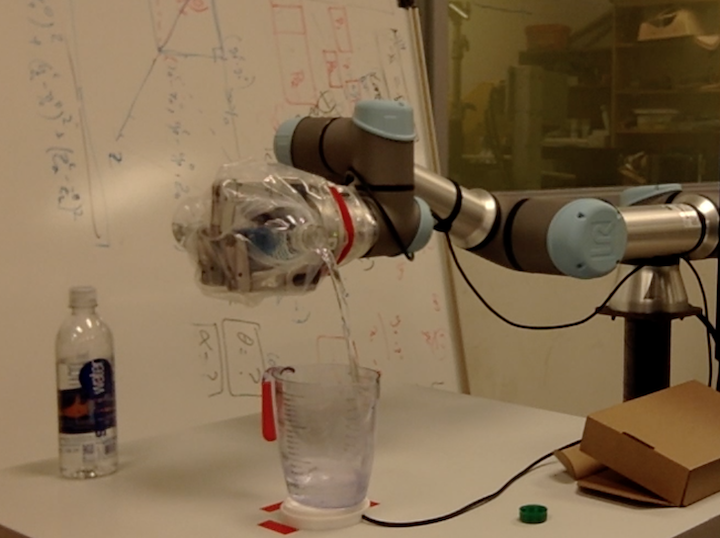}
\end{center}
\caption{Evaluating the model on UR5e collaborative robotic arm}
\label{fig-ur5e}
\end{figure}

\begin{figure*}[h]
\includegraphics[width=\linewidth]{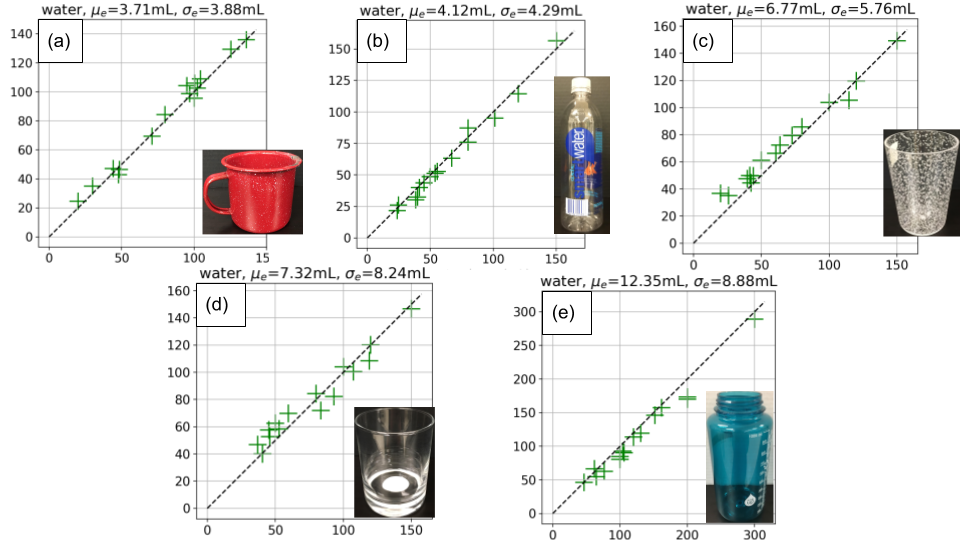}
\caption{Actual-vs-target comparison of pouring water using (a) red cup which is used for training (b) water bottle (c) cup with bubble pattern which we referred to as the bubble cup (d) glass cup (e) fat bottle.  }
\label{fig-water_accuracy}
\end{figure*}

We started with the task that has the lowest difficulty and tested the model by pouring water from the red cup that has been used for training. Figure \ref{fig-water_accuracy} (a) shows a small error of $\mu_e=3.71$mL, indicating that the learning is successful.
Then we increase the difficulty of the tasks and test the model by pouring water from different source containers that have not been used for training. Table \ref{table-error} summarizes the mean and standard deviation of the errors, $\mu_e$ and $\sigma_e$, in milliliters of the model pouring water from different source containers, and of the human pouring water from the red cup. Figures \ref{fig-water_accuracy} (b) through (e) show the error of five source containers whose $\sigma_e$ is smaller than that of human. Compared with the error of using the red cup $\mu_e=3.71$mL, the error of using the five source containers is larger, ranging from $\mu_e=4.12$mL to $\mu_e=12.35$mL, which is expected. Based on the results, we call them "accustomed containers". We have also evaluated the model on a UR5e robotic arm using the water bottle (Figure \ref{fig-ur5e}) in which $\mu_e$ = 7.83mL and $\sigma_e$ = 6.62mL.  

\begin{table}[h!]
\begin{center}
\begin{tabular}{| c | c | c | c |}
\hline
cup & \thead{cup \\ in training} & $\mu_e$ (mL) & $\sigma_e$ (mL) \\
\hline
red & yes & 3.71 & 3.88\\

water bottle & no & 4.12 & 4.29 \\

bubble & no & 6.77 & 5.76\\

glass & no & 7.32 & 8.24 \\

fat bottle & no & 12.35 & 8.88 \\

red (by human) & n/a & 12.37 & 9.80\\

measuring cup & no & 11.29 & 12.82 \\

wine bottle & no & 51.22 & 39.61 \\

blue bottle & no & 55.84 & 47.26 \\

\hline
\end{tabular}
\end{center}
\caption{Errors of pouring water from different source containers}
\label{table-error}
\end{table}             

\begin{figure}[h]
\centering
\includegraphics[width=0.7\linewidth]{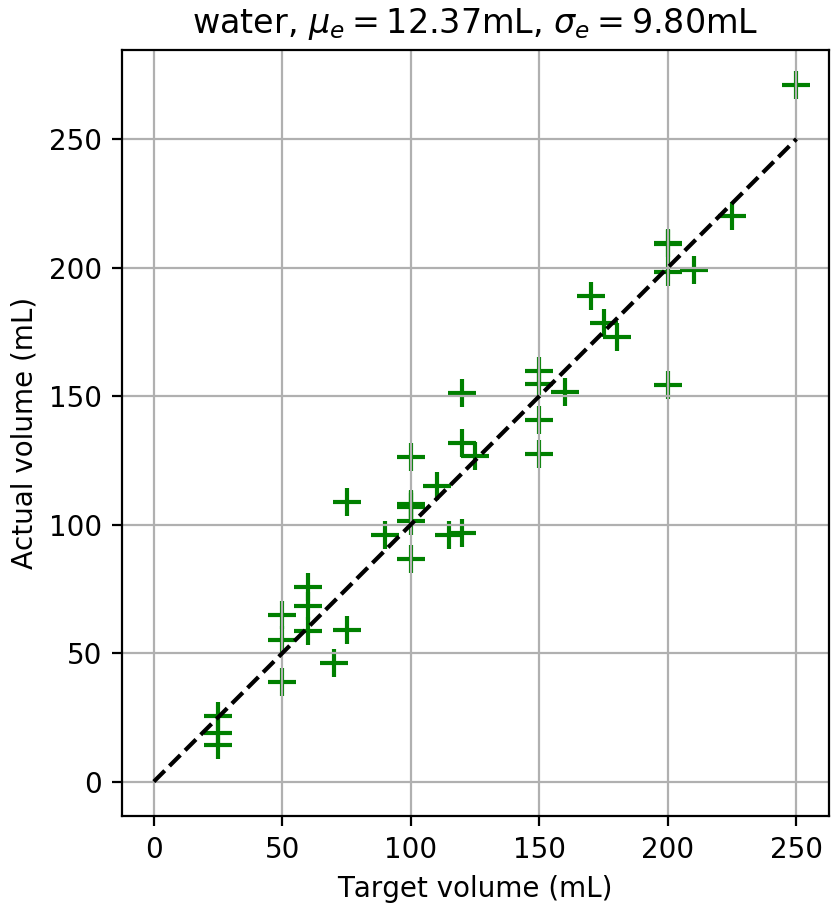}
\includegraphics[width=0.2\linewidth]{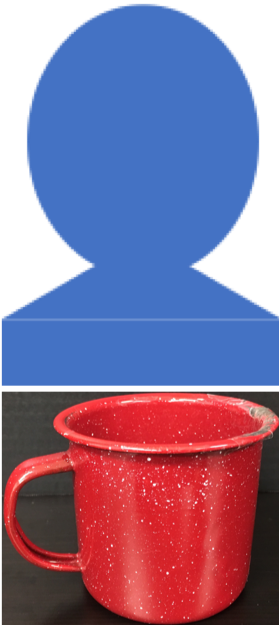}
\caption{Actual-vs-target comparison of 4 human subjects pouring water from the red cup}
\label{fig-human}
\end{figure}

We wanted to compare the pouring model with humans and therefore we asked four human subjects to do accurate water pouring with the red cup. We made an animation on a computer screen that shows the target volume and the real-time volume of water that has already been poured. The animation faithfully shows the fluctuation of the volume reading while pouring. The subjects were asked to look only at the animation and pour the target volume. They were asked to pour naturally and with a single pour. Pouring too fast or too slow was not allowed. We collected 10 trials with each subject, resulting in 40 total trials. Figure \ref{fig-human} shows the results of human accurate pouring: $\mu_e=12.37$mL and $\sigma_e=9.80$mL. Compared with humans, pouring water from the red cup (Figure \ref{fig-water_accuracy} (a)) achieves a lower $\mu_e=3.71$mL and $\sigma_e=3.88$mL. The model achieves lower error than humans because the model is trained using the actual outcomes.  

\begin{figure}
\centering
\includegraphics[width=0.95\columnwidth]{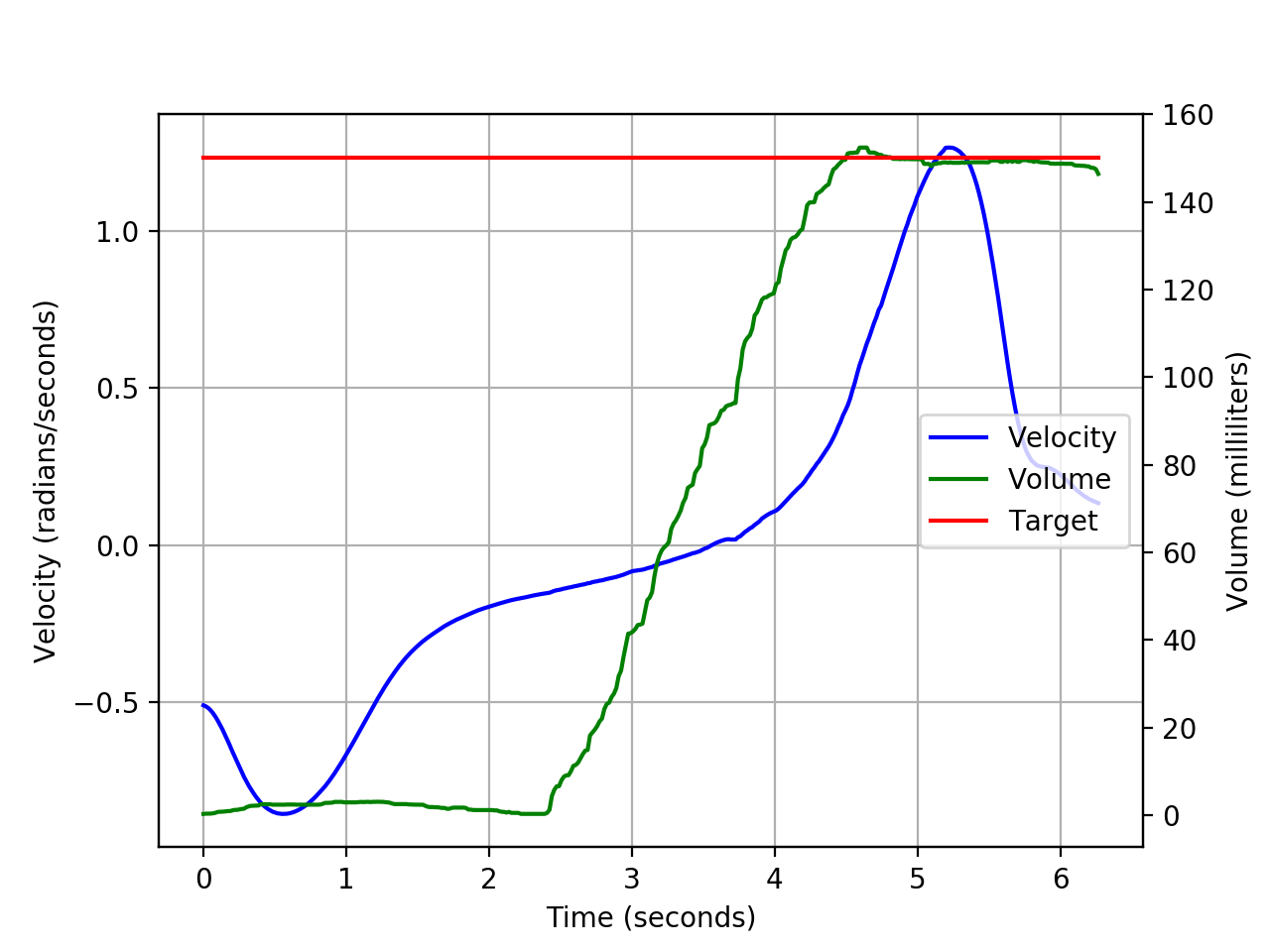}
\caption{An example of the robot pouring water using the water bottle. Velocity: velocity of the source container measured counter clock-wise, negative radians/seconds means that the container is going forward and positive means backward. Volume: volume of water in the receiving container. Target: target volume for the trial.}
\label{fig-robot-pouring}
\end{figure}

Figure \ref{fig-robot-pouring} shows an example of the resulting $\omega(t)$, $vol(t)$, and $vol_{2pour}$ for a trial of pouring water using the trained model with the water bottle. The target $vol_{2pour}$ was established as 150\,mL from an initial $vol_{total}$ of approximately 430\,mL. The final amount poured was 149\,mL, an error of 1\,mL. We can see that for this example the pouring was finished in less than 6 seconds and the water stopped to be poured out in less than 5 seconds. The small spike seen in the volume is due to the force sensor's noise generated by the water movement. We measured the final volume poured once the water had stabilized in the receiving container.

We have also tested the model on two containers that are significantly different from the training set: a wine bottle and a blue bottle that were available in our laboratory. As shown in Table \ref{table-error}, their mean volume errors were 51.22\,mL and 57.06\,mL, respectively. The mean volume error for the wine bottle and the blue bottle is 13 times higher than that of the red cup (3.71\,mL). They are unaccustomed containers to our pouring model because of their unusual shapes. 

\begin{figure}[h]
\centering
\includegraphics[width=0.95\columnwidth]{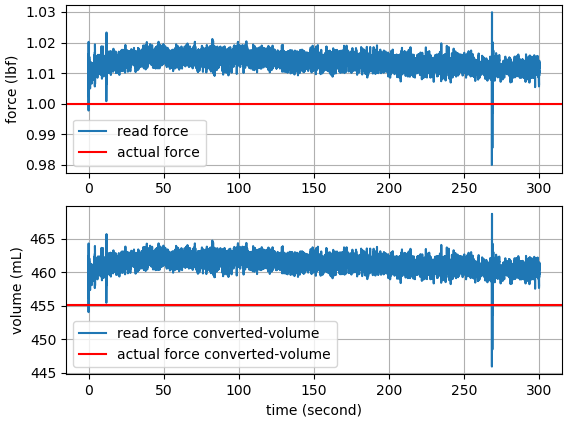}
\caption{Readings of the force sensor for a weight of 1.0 lbf taken during 300 seconds. The bottom subfigure shows the volume converted from force.}
\label{fig-sensor_drift}
\end{figure}

Having evaluated the error of the model pouring different but relatively large amounts of water, we evaluated the error of the model pouring a small amount of water. We use the model to pour 20mL and 15mL using the red cup, respectively, each for 15 times. For 20mL $\mu_e$ = 9.68mL and $\sigma_e$ = 7.96mL. For 15mL $\mu_e$ = 2.83mL and $\sigma_e$ = 3.33mL. Both $\mu_e$ and $\sigma_e$ for pouring 20mL are smaller than those of pouring a larger volume with the red cup (Figure \ref{fig-water_accuracy} (a)). The error of pouring 15mL is larger than both the error of pouring 20mL and the error of pouring a larger volume. 
In Figure \ref{fig-sensor_drift}, we plot the reading of the force sensor for a 1.0-lbf weight for 300 seconds. Figure \ref{fig-sensor_drift} also shows the water volume converted from the corresponding force. For a 1.0-lbf weight, the force sensor has a nonlinearity error of around 0.01 lbf, which is 1\% of 1.0 lbf. The corresponding error in volume is around 5mL.  

\subsubsection{Comparison with Related Works}

The difficulty of accurate pouring increases as the duration it takes to pour decreases. Our model pours as fast as humans. The duration of each pour in the human demonstrations dataset ranges from 3.2 to 8.7. The duration range of our model was 2.8 to 7.6 seconds. In comparison, \cite{7989307} achieves 38\,mL error using 25 seconds for each pour using similar containers as our accustomed containers, it takes much longer than ours and causes a larger error. \cite{8653969} achieves under 5mL error and uses 20-45 seconds for each pour, it achieves lower error than ours but takes longer than \cite{7989307}. We are not aware of any prior art that has evaluated containers as diverse as ours.

Our approach uses weight to monitor the poured volume. \cite{7989307, do2018accurate, chaudo2018} also use one single modality (weight or vision) to monitor the poured volume, whose reported pouring error is 38mL \cite{7989307}, 23.9mL, 13.2mL, and 30.5mL \cite{do2018accurate}, and 19.96mL \cite{chaudo2018} respectively. \cite{8653969} achieves an error that is under 5mL, but two modalities (weight and vision) are used to monitor the poured volume. It also pours slowly, a behavior that can help with the higher accuracy achieved. The error reached by the model lies between 3.71mL and 12.35mL, lower than the above approaches.

In our previous work \cite{tianze2019}, we used model predictive control (MPC) to address the problem of accurate pouring. The proposed controller uses RNN to predict the weight of the liquid in the receiving container and controls the angular velocity of the source container. We evaluated the performance of the controller by comparing it with a switch controller. The switch controller applies a constant forward velocity to the source container when the volume in the destination container is less than the target. It applies a constant backward velocity when the volume reaches the target. Table \ref{tab-mpc} shows the results. The Switch $\omega_1$ controller used $20$ deg/sec as the forward velocity and $-30$ deg/sec as the backward velocity. The Switch $\omega_2$ controller used $5$ deg/sec as the forward velocity and $-7.5$ deg/sec as the backward velocity. We can see that when the forward angular velocity becomes smaller, the mean volume error decreases. This result is expected since the difficulty of controlling the volume in the destination container decreased. Based on Table \ref{tab-mpc}, the errors for the MPC controller go from 7.25mL and 26.13mL. We can see that model $M_0$ also performs better than the MPC controller. Without the guarantee that the RNN-generated physics model is highly precise, the performance of the powerful MPC algorithm is compromised. Pouring is not a trivial task that can be solved by a traditionally powerful algorithm.

\begin{table}[h]
\begin{center}
\begin{tabular}{|c|c|c|c|c|}
\hline
cup  & model & \thead{cup \\ in training} & $\mu_e$ (mL) & $\sigma_e$ (mL) \\
\hline
\multirow{3}{*}{red} & MPC & \multirow{3}{*}{yes} & 7.25 & 4.92\\
 & Switch $\omega_1$ &  & 33.50 & 7.76\\
 & Switch $\omega_2$ &  & 4.50 & 1.87\\
\hline
glass & MPC & no & 14.25 & 9.11\\
\hline
bottle & MPC & no & 15.88 & 5.13\\
\hline
fat & MPC & no & 18.25 & 8.30\\
\hline
\multirow{3}{*}{bubble} & MPC & \multirow{3}{*}{no} & 26.13 & 6.29\\
 & Switch ($\omega_1$) &  & 56.25 & 5.85\\
 & Switch ($\omega_2$) &  & 22.25 & 4.29\\

\hline
\end{tabular}
\end{center}
\caption{Results of pouring with MPC or Switch Controller}
\label{tab-mpc}
\end{table}   

\subsubsection{Model Evaluation on Different Materials} \label{sect:eval_solid}
\begin{figure}
\centering
\includegraphics[width=0.95\columnwidth]{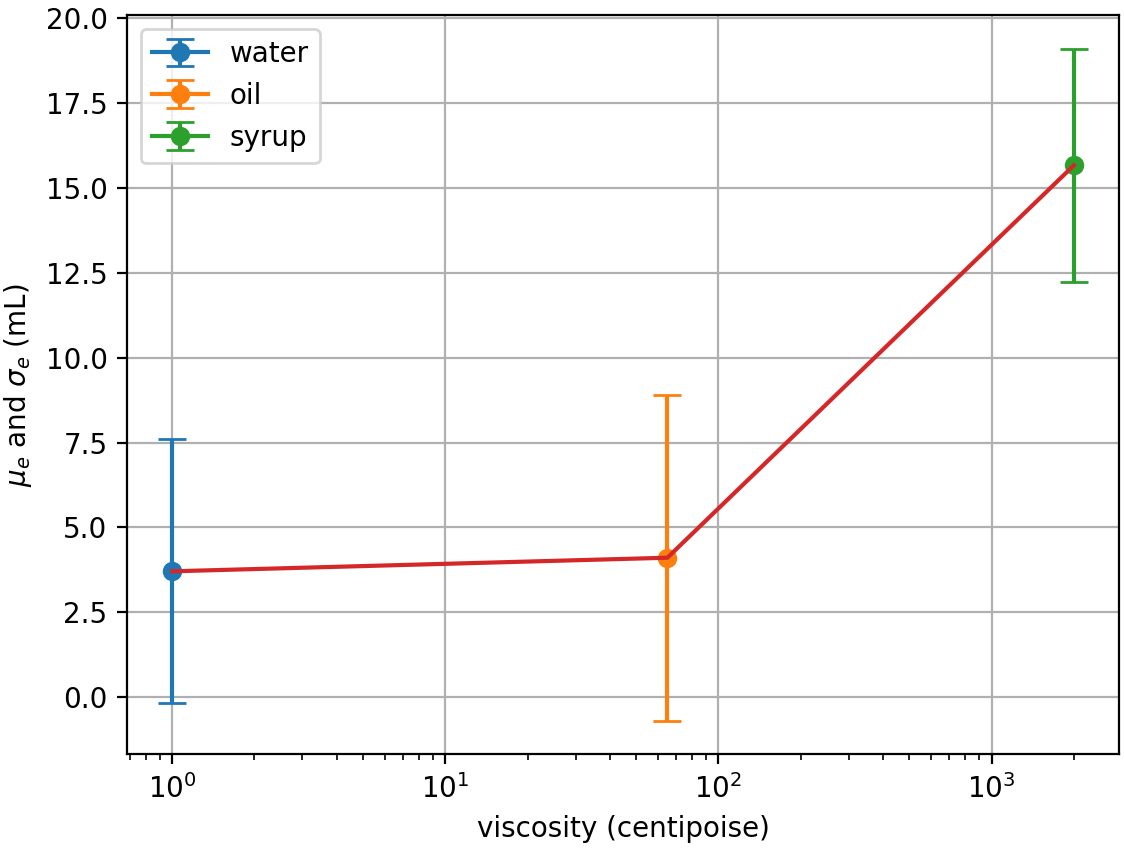}
\caption{Pouring accuracy of liquids with different viscosity. x-axis plotted in logarithmic scale.}
\label{fig-viscosity}
\end{figure}

We also tested the model on liquids with different viscosity from water such as cooking oil and syrup. We used the red cup to pour since it is part of the training dataset. However, the dataset comes from pouring motions of water. 
We speculate that viscosity played an important role in the accuracy of pouring different kinds of liquids. Therefore, Figure \ref{fig-viscosity} shows the error bars of pouring water, oil, and syrup with the red cup versus their viscosity for 15 pours. The three types of liquids have very different viscosities. We use 1 centipoise (cps) as the viscosity for water, 65 centipoises for oil, and 2000 for syrup. We plot the viscosities on a logarithmic scale. We can see that the mean error increases as the viscosity increases. The relationship is neither linear nor exponential. 

\begin{figure}
\centering
\includegraphics[width=0.95\columnwidth]{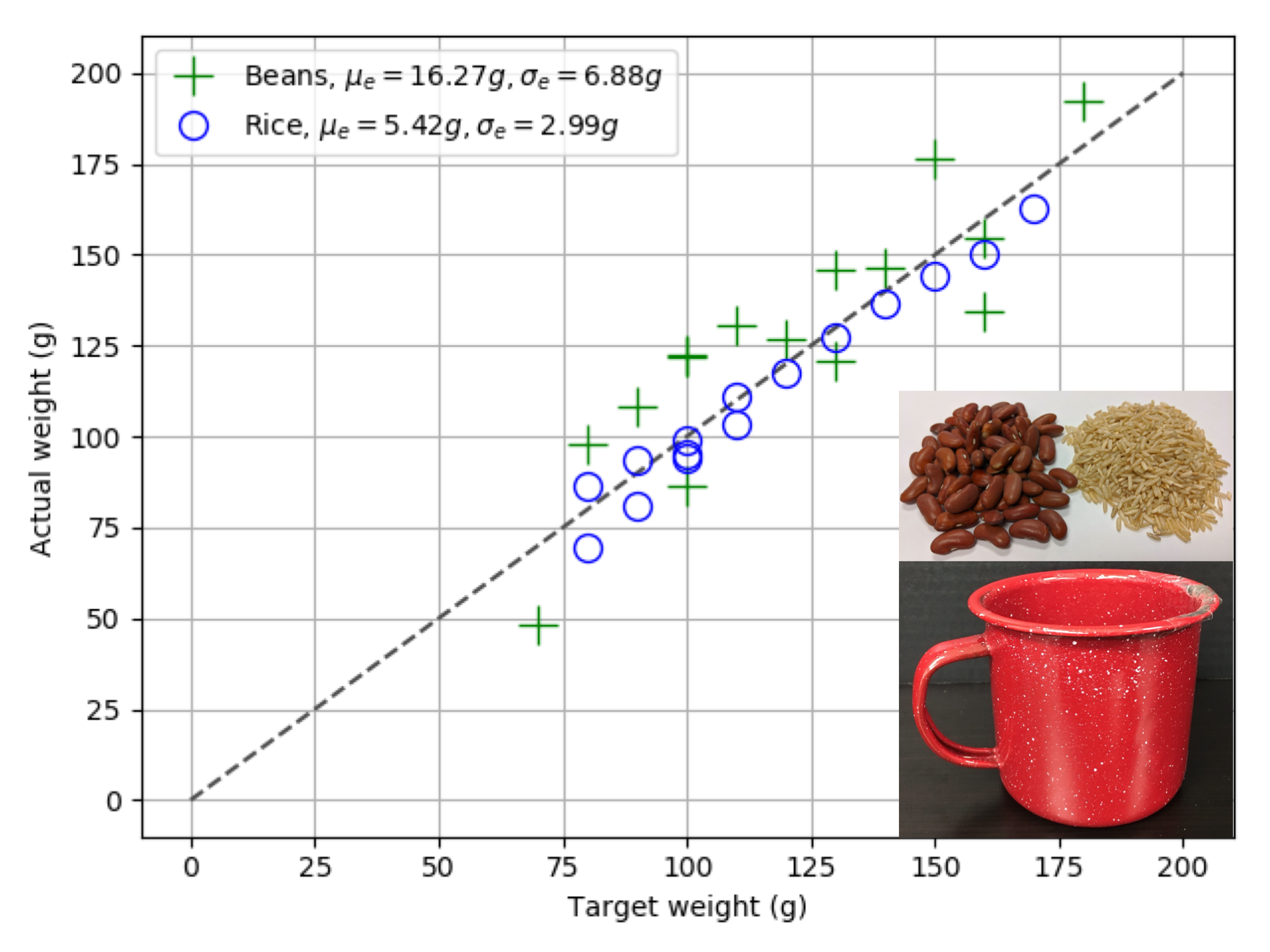}
\caption{Pouring accuracy of beans and rice using the red cup.}
\label{fig-solids}
\end{figure}

We also evaluated the model on granular materials poured in cooking scenarios: beans and rice. We used the same model trained from pouring water. Figure \ref{fig-solids} shows the result of 15 pours of beans and rice using the red cup. We changed the unit of measure to grams (g) as it is more suitable for solid materials than volume. We can see that the mean error is small for rice, whereas it is higher for beans. However, the mean errors presented by the model are similar to those presented in \cite{solid2019pouring}, where the authors state to have a mean error estimation of 14.3g for red beans and 4.36g for pouring rice. Their approach is not meant for accurate pouring but for pouring mass estimation based on fingertip sensors of a robotic hand.

\subsection{GSSP Evaluation -- Generalization to unaccustomed containers}
\label{sect:evalcon}

\begin{figure}[h!]
    \centering
    \includegraphics[width=0.95\columnwidth]{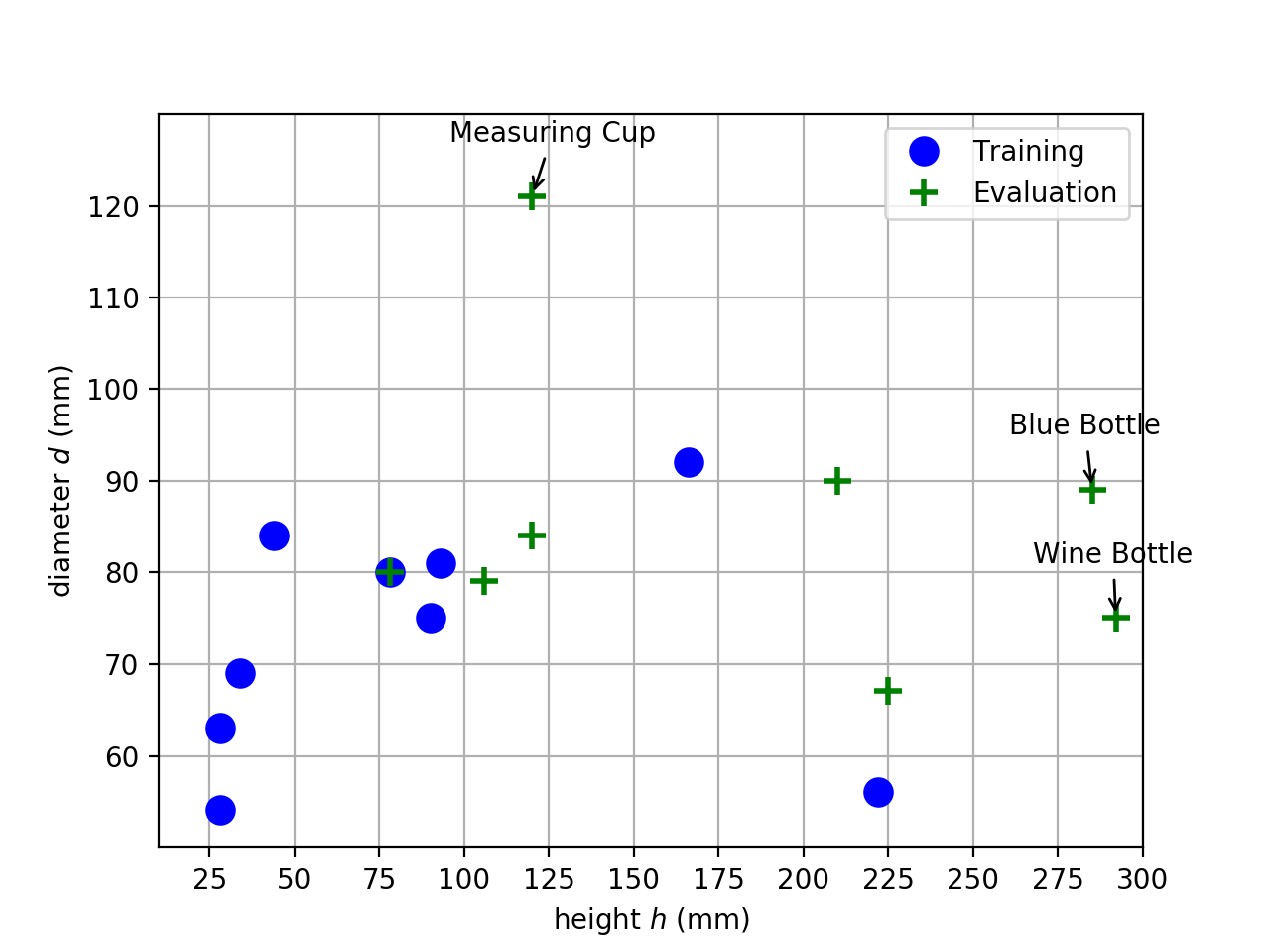}
    \caption{Scatter plot of height vs diameter for source containers used for experiments.}
\label{fig-cups-distr}
\end{figure}

We evaluated GSSP on the two unaccustomed containers -- a wine bottle and a blue bottle.  For comparison, we also added the measuring cup into the unaccustomed container set.
We chose those three containers because, as Table \ref{table-error} shows, the wine bottle and blue bottle had the highest mean errors among all the test containers evaluated. Although the measuring cup did not present a significantly high mean error, it did present a much higher standard deviation error than humans. Figure \ref{fig-cups-distr} shows the scatter plot of the height versus diameter of the source containers used for experiments. We can see that the wine bottle and the blue bottle are both much taller than the rest of the containers, and the measuring cup has a much larger diameter.

We evaluated GSSP using batch fine-tuning for the wine bottle, blue bottle, and measuring cup. We evaluated the accuracy of the resulting models by pouring water 15 times per experiment for batch fine-tuning and carried out the experiments maintaining the same set of volumes for a fair comparison. We also evaluated GSSP using gradual fine-tuning for the wine bottle. We refer to the initially learned model that was evaluated in Section \ref{sect:eval1} as $M_0$. 

\subsubsection{Wine Bottle} \label{sec:wine}
\begin{figure}
    \centering
    \includegraphics[width=0.95\columnwidth,trim=2em 3em 2em 0em, clip]{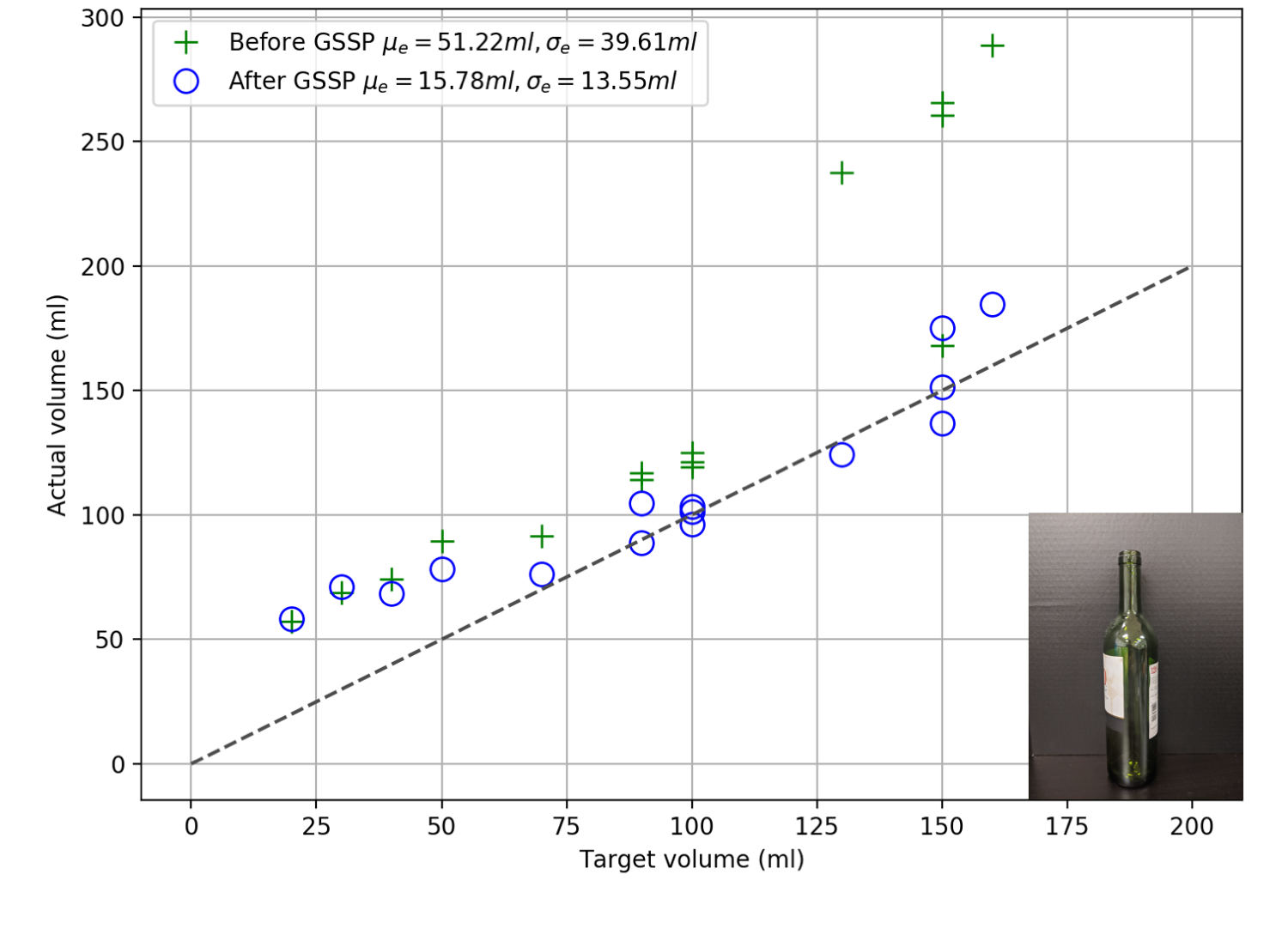}
    \caption{Results for Wine Bottle before and after GSSP.}
    \label{fig-wine-cup-m0}
\end{figure}

We executed a total of 36 practices using the initial model $M_0$ with the wine bottle for different $f_{total}$ and $f_{2pour}$. Then, we fine-tuned the model using the obtained dataset. We call the fine-tuned model $M_1$.
Figure \ref{fig-wine-cup-m0} shows the mean and standard deviation of the error for 15 pours before and after applying GSSP, i.e., using models $M_0$ and $M_1$, respectively. We can see that the wine bottle's mean volume error became 15.78\,mL, a reduction of around 69\% from 51.22\,mL mean error. After applying GSSP, some trials over pour but others under pour. When using model $M_0$, all trials over pour water. 

We also tested the gradual fine-tuning approach using the wine bottle for which we set the number of practices $n=10$. The set of volume variations $\mathcal{R} = \{(f_{total}^{i}, f_{2pour}^{i})\}$ for $i=1,...,10$, was chosen to be the same for each iteration of the algorithm. Table \ref{tab-gradual-wine} shows the accuracy evolution of the fine-tuning algorithms we carried out. The mean and standard deviation errors from the table's first row are different from the ones shown in Table \ref{table-error} as the set $\mathcal{R}$ and the number of trials was different for both experiments. We hypothesize that running more iterations of the algorithm will further improve the result. However, we believe that there should exist enough variation in the selection of $f_{total}$ and $f_{2pour}$ to outperform the result of batch fine-tuning for this particular container.

\begin{table}[h!]
    \centering
    \begin{tabular}{| c | c | c | c |}
\hline
 \thead{Base \\ Model} & \thead{Fine-tuned \\ Model} & $\mu_e$ (mL) & $\sigma_e$ (mL) \\
\hline
  $M_0$ & &  80.23 & 49.13\\
  $M_0$ & $M_{2}$  &  38.67 & 11.98\\
  $M_{2}$ & $M_{3}$  &  30.04 & 17.26\\
  $M_{3}$ & $M_{4}$  &  18.21 & 8.76\\
\hline 
\end{tabular}
\caption{Accuracy for Wine Bottle after gradual fine-tuning.}
    \label{tab-gradual-wine}
\end{table}

Comparing the results of Table \ref{tab-gradual-wine}'s fourth row with Figure \ref{fig-wine-cup-m0} after applying GSSP, we can see that both methodologies yield similar results. Gradual fine-tuning has the advantage over batch fine-tuning w.r.t. the cost it takes to collect the practices. However, there exists a trade-off in training time: batch fine-tuning only trains once, while gradual fine-tuning trains several times. Nevertheless, gradual fine-tuning allows us to realize whether there is an improvement or not after the first iteration of the algorithm using only a few practices. In batch fine-tuning, after taking a considerable time carrying out practices, we expect that there exists an improvement, but the practices collected may lead to unsatisfactory results. 
\subsubsection{Blue Bottle} \label{sec:blue}
\begin{figure}
    \centering
    \includegraphics[width=0.95\columnwidth,trim=2em 2em 2em 0em, clip]{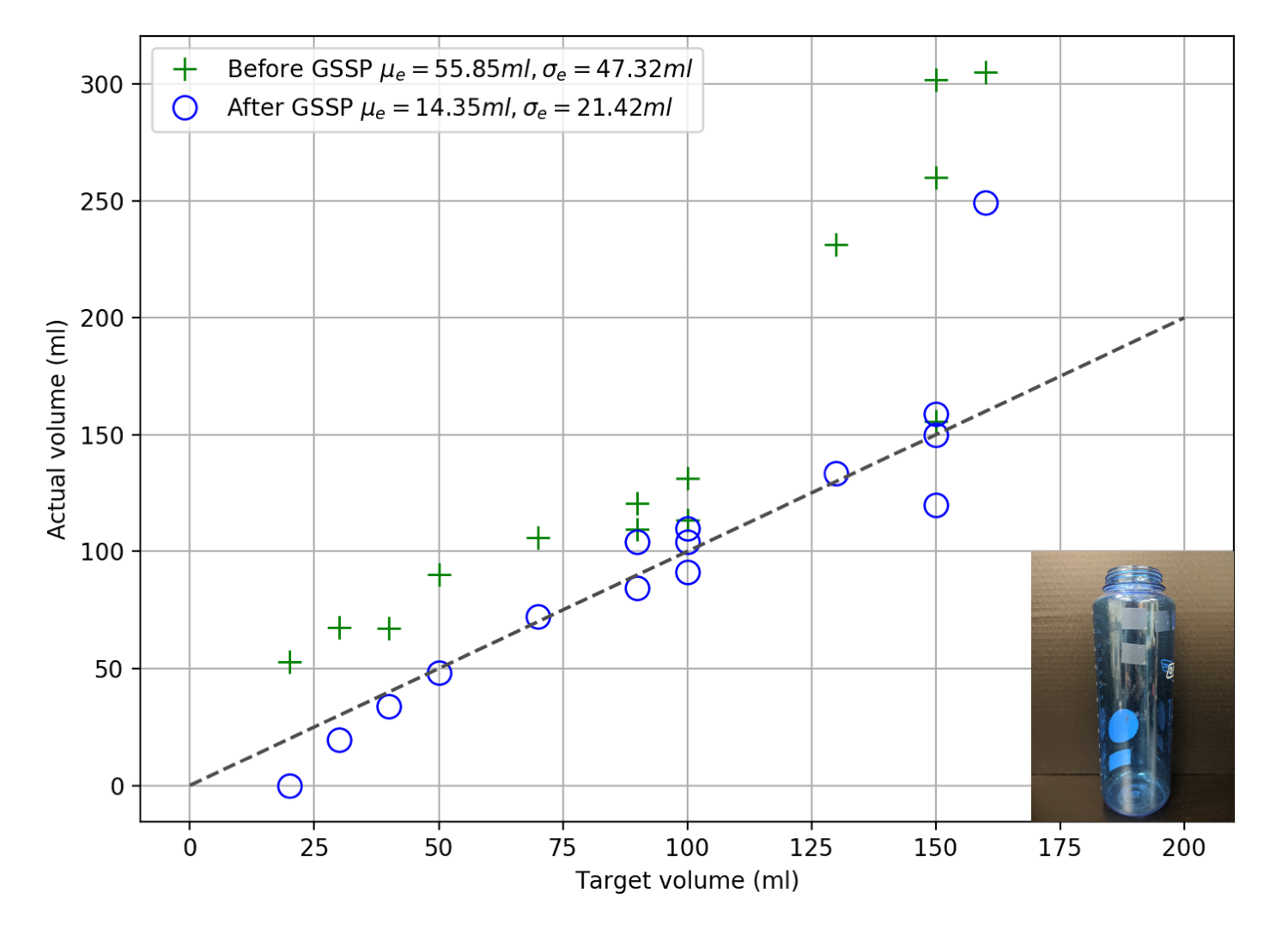}
    \caption{Results for Blue Bottle before and after GSSP.}
    \label{fig-blue-cup-m0}
\end{figure}

We decided to apply only batch fine-tuning for the blue bottle. We executed 54 practices using the blue bottle and fine-tuned $M_0$. We call the resulting model $M_5$. Figure \ref{fig-blue-cup-m0} shows the mean and standard deviation of the error for 15 pours before and after applying GSSP, i.e., using $M_0$ and $M_5$, respectively. We can see a reduction of 74\% in average error from 55.85\,mL to 14.35\,mL. We can also see that there exists an outlier for the target of 160\,mL that may be affecting the standard deviation. However, there was a 54\% reduction in this statistic from 47.32\,mL to 21.42\,mL.

\subsubsection{Measuring Cup}
\label{sec:measuring}
\begin{figure}
    \centering
    \includegraphics[width=0.95\columnwidth,trim=2em 2em 2em 0em, clip]{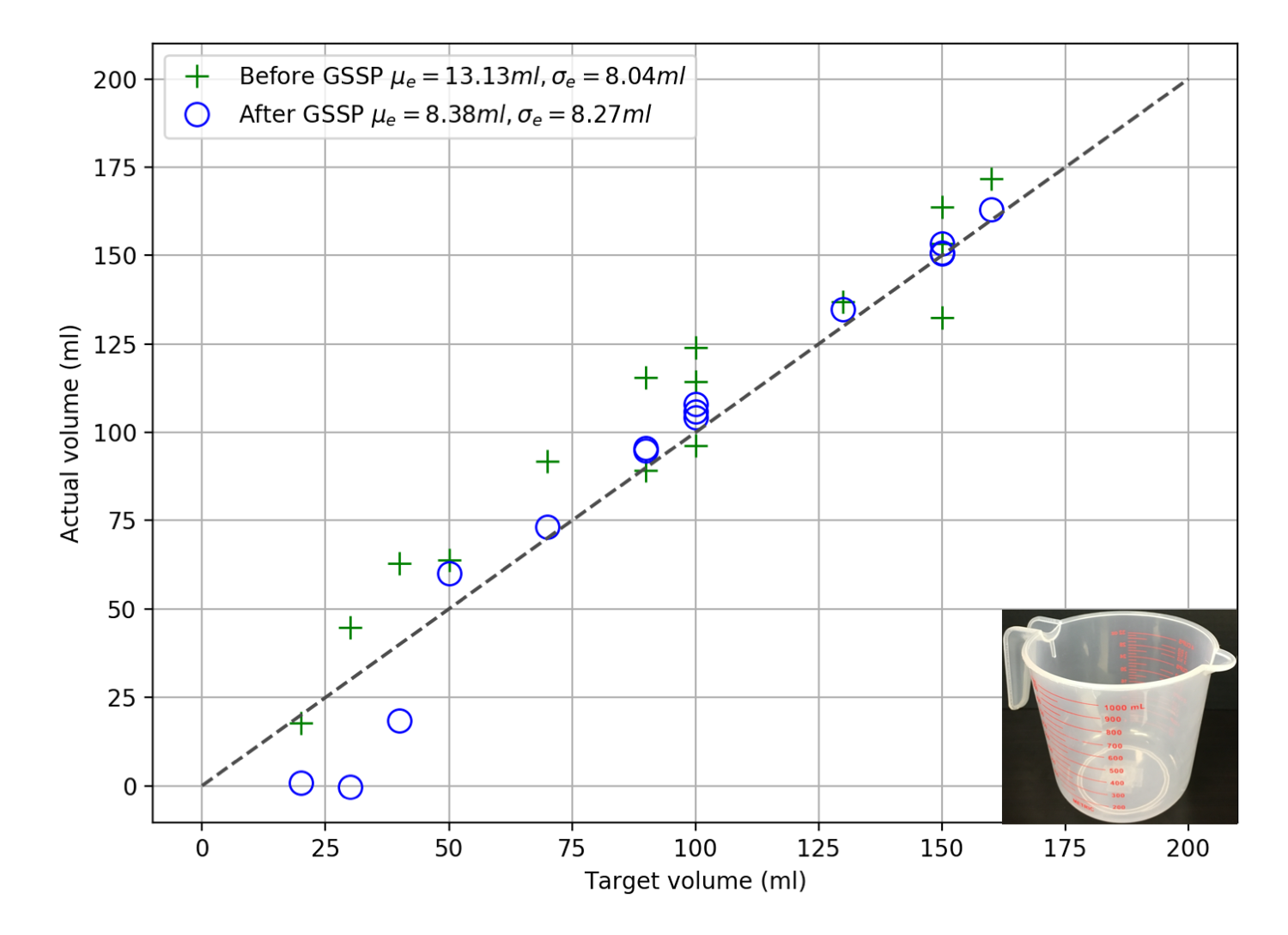}
    \caption{Results for Measuring Cup before and after GSSP.}
    \label{fig-fat-cup-m0}
\end{figure}

We executed 36 practices using the measuring cup. We also decided to use batch fine-tuning for this source container and applied GSSP. We call the resulting model $M_6$.
Figure \ref{fig-fat-cup-m0} shows the scatter plot of the target versus the actual poured volume for the measuring cup before and after GSSP, i.e., when using $M_0$ and $M_{6}$, respectively. 
We can see a reduction again in mean error when comparing the target with the actual volume poured by the robot. 

\subsection{GSSP Evaluation -- Generalization to new material} \label{sect:evalmat}

We have also carried out GSSP experiments on syrup and red beans. We used the red cup as the container for pouring with the model $M_0$. This model comes from training using the human demonstrations dataset of pouring water. We selected the red cup as it is the most accurate container for pouring water with model $M_0$.

\subsubsection{Syrup} 
\begin{figure}
    \centering
    \includegraphics[width=0.95\columnwidth]{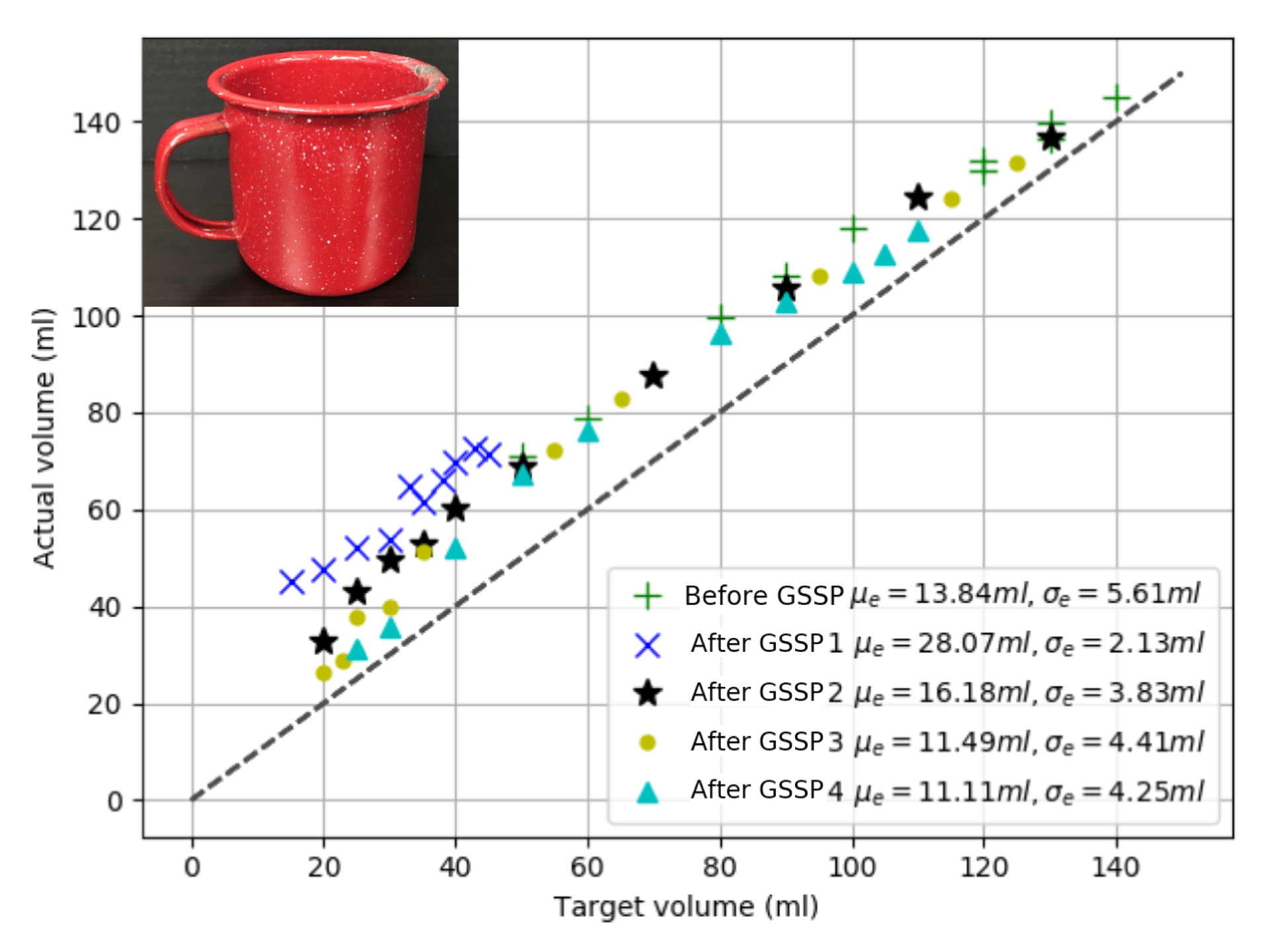}
    \caption{Accuracy for syrup before and after 4 gradual fine-tuning iterations.}
    \label{fig-gip-syrup}
\end{figure}

We applied gradual fine-tuning for syrup using model $M_0$ for which we set the number of practices $n=10$. Figure \ref{fig-gip-syrup} shows the evolution of accuracy for 5 iterations of Algorithm \ref{alg:gip}. We can see that the initial mean error of 13.84\,mL was mostly affected for target volumes lower than 100\,mL. For the second iteration, we decided to pour small volumes for which we could see a mean volume error of 28.07\,mL. For the third iteration, we could see a reduction of the mean error from 28.07\,mL to 16.18\,mL (42\% reduction). We could see that the improvement was caused by targets lower than 50\,mL and higher than 80\,mL. For the fourth iteration, we could see an improvement from 16.18mL to 11.49\,mL (28\% reduction). At this stage, we were able to reduce the initial mean error from 13.8\,4mL to 11.49\,mL (17\% reduction). Finally, we carried out a fifth iteration for which the mean error decreases from 11.49\,mL to 11.11\,mL (3\% reduction).

\subsubsection{Red Beans} 
We collected a total of 36 practices for pouring beans with different initial target weights, where the 15 pouring trials shown in Figure \ref{fig-solids} are included. Table \ref{tab-batch-beans} summarizes the mean and standard deviation weight errors. Model  $M_{11}$ results from fine-tuning $M_0$ using the 36 pouring trials collected. 
We can see that GSSP not only works for reducing the error when pouring liquids from new source containers. It also can be used to improve the accuracy of pouring a different material. Interestingly, the model $M_0$ trained for pouring water has a considerable small error when used for pouring beans.

\begin{table}[h!]
\centering
\begin{tabular}{| c | c | c | c | c | }
\hline
\thead{Source \\ Container} & \thead{Model}  & $\mu_e$ (g) & $\sigma_e$ (g) \\
\hline
\multirow{2}{*}{Red Cup} &  $M_{0}$  &  16.27 & 6.88\\

 & $M_{11}$  &  11.49 & 6.98\\

\hline
\end{tabular}
\caption{Comparison of accuracy for Red Cup before and after fine-tuning $M_0$ using data collected for red beans.}
\label{tab-batch-beans}
\end{table}

\subsection{GSSP Discussion}

\begin{figure*}[h!]
\resizebox{\linewidth}{!}{
    \begin{forest}
for tree={
  minimum height=1cm,
  anchor=north,
  align=center,
  child anchor=north,
  edge={-stealth,line width=1pt},
},
[{$M_0$}, align=center, name=BM
  [{Practices}, name=DS
    [{Gradual}, name=GT
    [{Wine \\ Bottle}, name=CM
        [{$M_{2\textnormal{-}4}$}, name=RM]
    ]
    [{Syrup} [{$M_{7\textnormal{-}10}$}]]
    ]
    [{Batch}
    [{Wine \\ Bottle} [{$M_1$}]]
    [{Blue \\ Bottle} [{$M_5$}]]
    [{Measuring \\ Cup} [{$M_6$}]]
    [{Red \\ Beans}[{$M_{11}$}]]
    ]
  ]
  [{Practices + \\ Human Demonstrations}
  [{Batch}[{Wine \\ Bottle} [{$M_{12}*$}]]
    [{Blue \\ Bottle} [{$M_{13}$}]]
    [{Measuring \\ Cup} [{$M_{14}$}]]
    ]
  ]
]
\node[anchor=west,align=left] 
  at ([xshift=-2.5cm]RM.west|-RM) {Result \\ Model(s)};
\node[anchor=west,align=left] 
  at ([xshift=-2.5cm]RM.west|-CM) {Container/ \\ Material};
\node[anchor=west,align=left] 
  at ([xshift=-2.5cm]RM.west|-GT) {GSSP Type};
\node[anchor=west,align=left] 
  at ([xshift=-2.5cm]RM.west|-DS) {Dataset};
\node[anchor=west,align=left] 
  at ([xshift=-2.5cm]RM.west|-BM) {Base \\ Model};
\end{forest}
\begin{forest}
for tree={
  minimum height=1cm,
  anchor=north,
  align=center,
  child anchor=north,
  edge={-stealth,line width=1pt},
},
[{$M_{12}*$}, align=center, name=BM
  [{Practices}, name=DS
    [{Batch}, name=GT
    [{Blue \\ Bottle}, name=CM
        [{$M_{15}$}, name=RM]
    ]
  ]
  ]
  [{Practices + \\ Human Demonstrations}
  [{Batch}[{Blue \\ Bottle} [{$M_{16}$}]]
  ]
]
]
\end{forest}
}
    \caption{Model relationships for the application of batch or gradual GSSP to particular datasets, containers, or materials.}
    \label{fig-model-relationships}
\end{figure*}
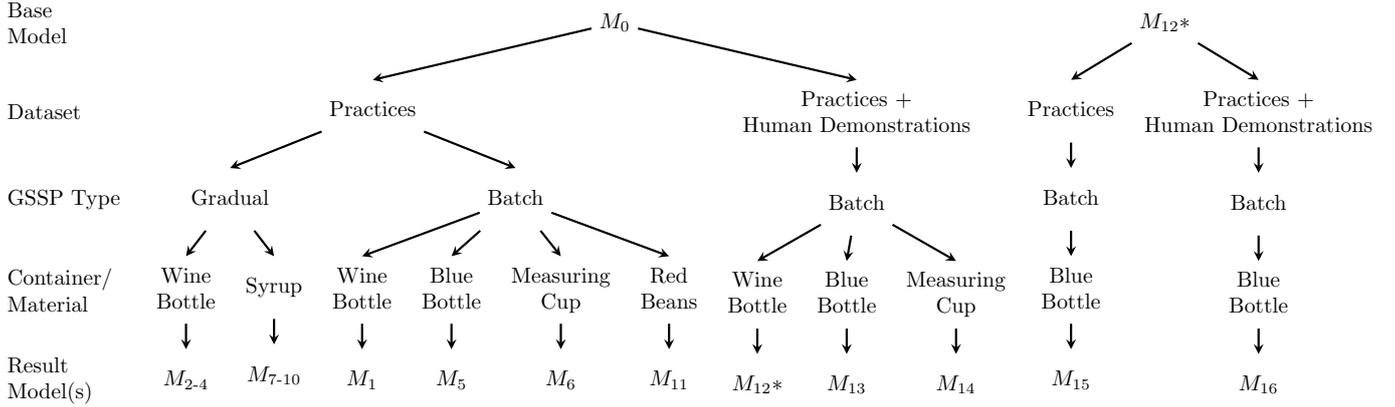

At this point, we have applied GSSP as formulated in Algorithm \ref{alg:gip} where the dataset used to fine-tune comes entirely from the robot practices. We performed further experiments to analyze the effects of combining the human demonstrations dataset described in section \ref{sec:dataset} with the robot practices. The hierarchical diagram of Figure \ref{fig-model-relationships} illustrates the details of the fine-tuned models that we have presented so far and the new ones we will present next. 
The models presented in sections \ref{sect:evalcon} and \ref{sect:evalmat} correspond to the children of the ``Practices" branch of Figure \ref{fig-model-relationships}. Such models were derived from fine-tuning $M_0$ with robot practices using the same model. 
In the following sections, we will compare the results of new models fine-tuned with a combination of robot practices and human demonstrations, versus the models fine-tuned with robot practices only. We also carried out experiments using the red cup with the fine-tuned models to verify the impact of accuracy to the accustomed containers. Finally, we will also show the results of applying GSSP to the robot practices using a fine-tuned model instead of $M_0$.

\subsubsection{Combination of Practices and Human Demonstrations}
We fine-tuned model $M_0$ using the combination of the human demonstrations dataset with the robot practices. We believe this is the same as training from scratch using the combined datasets, the only difference being that the training converges faster as $M_0$ already learned from the human demonstrations. Table \ref{tab-summary-comb} shows the summary of mean and standard deviation errors for the fine-tuned models. $M_{12}$ is the fine-tuning of $M_0$ with the training dataset of human demonstrations from section \ref{sec:dataset} plus the batch of wine bottle robot practices from section \ref{sec:wine}. Similarly, $M_{13}$ and $M_{14}$ are the results of fine-tuning $M_0$ with the training dataset of human demonstrations of section \ref{sec:dataset} plus the blue bottle and measuring cup batches of robot practices, respectively. 

Model $M_{12}$ has slightly higher mean and standard deviation errors than model $M_1$ w.r.t. wine bottle, meaning that model $M_1$ is better suitable for pouring accurately with the wine bottle. Interestingly, model $M_{13}$ slightly outperformed model $M_5$  w.r.t. the blue bottle's accuracy. Similarly, model $M_{14}$ marginally outperformed model $M_6$. Given this marginal improvement, fine-tuning using only the robot practices is sufficient to achieve a satisfactory precision. Moreover, the original human demonstrations dataset is necessary to generate models $M_{13}$ and $M_{14}$ with the increased cost of training with a larger dataset.

\begin{table}[h!]
    \centering
    \begin{threeparttable}
    \begin{tabular}{| c | c | c | c | c | c |}
\hline
\thead{Source \\ Container} &
 \thead{Base \\ Model} &  \thead{Fine-tuned \\ Model} & $\mu_e$ (mL) & $\sigma_e$ (mL) \\
\hline
\multirow{2}{*}{\thead{Wine Bottle}} 
& \multirow{2}{*}{$M_{0}$} &   $M_{1}$\tnote{a}  &  15.78 & 13.55\\
&  &   $M_{12}$\tnote{b}  &  17.43 & 13.65\\

\hline
\multirow{2}{*}{\thead{Blue Bottle}}
& \multirow{2}{*}{$M_{0}$} &   $M_{5}$\tnote{a}  &  14.35 & 21.42\\
&  & $M_{13}$\tnote{b}  &  12.65 & 5.99\\
\hline
\multirow{2}{*}{\thead{Measuring \\ Cup}} 
& \multirow{2}{*}{$M_{0}$} & $M_{6}$\tnote{a}  &  8.38 & 8.27\\
&  & $M_{14}$\tnote{b}  &  8.03 & 8.34\\
\hline
\end{tabular}
\begin{tablenotes}
\footnotesize
  \item[a] Model fine-tuned with robot practices only.
  \item[b] Model fine-tuned with robot practices plus human demonstrations.
  \end{tablenotes}
  \end{threeparttable}
\caption{Accuracy comparison of pouring water using the fine-tuned models.}
    \label{tab-summary-comb}
\end{table}

\subsubsection{GSSP Effect on Accustomed Containers}
We investigated how the fine-tuned models affected the generalization of pouring containers that already work well with $M_0$. To this end, we selected the red cup as it was the best performing container for $M_0$ and is also part of the human demonstrations dataset. From Table \ref{table-error}, it has $\mu_e =$  3.71\,mL and $\sigma_e =$ 3.88\,mL. Table \ref{tab-summary-red} shows the accuracy of 15 pouring trials using the red cup using models fine-tuned with robot practices only and the combination of practices with human demonstrations. Based on the results, we can see that the accuracy of the red cup was severely affected by the fine-tuned models that come from the practices of the wine bottle and the blue bottle, i.e., $M_1$ and $M_5$, respectively. The accuracy was not drastically impacted by using $M_6$. 

We believe this is related to the fact that $M_0$ was already performing well with the measuring cup. Therefore, the fine-tuning needed to learn more, e.g., modifies $M_0$ more drastically, from the practices of the wine bottle and the blue bottle than from the measuring cup's. Based on this result, we can state that models $M_1$ and $M_5$ are specialized models that accurately pour with the wine bottle and the blue bottle, respectively. At this stage, the robot can use a selector such that when it needs to use the wine bottle, $M_1$ is chosen to pour. Similarly, when it needs to use the blue bottle, $M_5$ is chosen to pour.

\begin{table}[h!]
    \centering
    \begin{threeparttable}
    \begin{tabular}{| c | c | c | c | c | c |}
\hline
\thead{Source \\ Container} &
 \thead{Base \\ Model} &  \thead{Fine-tuned \\ Model} & $\mu_e$ (mL) & $\sigma_e$ (mL) \\
\hline
\multirow{6}{*}{\thead{Red Cup}} 
& \multirow{6}{*}{$M_{0}$} &   $M_{1}$\tnote{a}  &  31.88 & 19.60\\
&  &   $M_{12}$\tnote{b}  &  8.67 & 5.13\\

\cline{3-5}
&  &   $M_{5}$\tnote{a}  &  25.29 & 14.98\\
&  & $M_{13}$\tnote{b}  &  8.52 & 5.51\\
\cline{3-5}
&  & $M_{6}$\tnote{a}  &  4.84 & 2.79\\
&  & $M_{14}$\tnote{b}  &  6.95 & 5.57\\
\hline
\end{tabular}
\begin{tablenotes}
\footnotesize
  \item[a] Model fine-tuned with robot practices only.
  \item[b] Model fine-tuned with robot practices plus human demonstrations.
  \end{tablenotes}
  \end{threeparttable}
\caption{Accuracy of pouring water with the Red Cup using the fine-tuned models.}
    \label{tab-summary-red}
\end{table}

\subsubsection{GSSP on a Different Base Model}

We also investigated the effect of applying GSSP to different starting models. We selected model $M_{12}$ that was the result of fine-tuning model $M_0$ using the combination of the dataset of human demonstrations plus the wine bottle robot practices. We used the blue bottle to collect 54 practices using model $M_{12}$. Table \ref{tab-different-start-blue} summarizes the results. We can see that the blue bottle's accuracy using model $M_{12}$ resulted in $\mu_e =$  26.38\,mL and $\sigma_e =$ 38.09\,mL. This result outperforms its counterpart when using $M_0$, where $\mu_e =$  55.84\,mL and $\sigma_e =$ 47.26\,mL. Model $M_{12}$ is more accurate to pour with the blue bottle than $M_0$. We fine-tuned model $M_{12}$ using only the blue bottle's practices, which resulted in model $M_{15}$. Its improvement was marginal around 4.7\% (25.12\,mL from 26.38\,mL). We also used the combination of the human demonstrations plus the wine bottle practices plus the blue bottle practices to fine-tune $M_{12}$. This resulted in model $M_{16}$. We could see that there was no improvement in the average error. Therefore, we believe that the most reliable model to fine-tune is the one that comes from human demonstrations, i.e., $M_{0}$. We hypothesize that such model has learned the variations inherent to humans. Therefore, it has more information than a fine-tuned model with the robot practices.

\begin{table}[h!]
\centering
\begin{threeparttable}
\begin{tabular}{| c | c | c | c | c | }
\hline
\thead{Source \\ Container} & \thead{Base \\ Model} & \thead{Fine-tuned \\ Model} & $\mu_e$ (mL) & $\sigma_e$ (mL) \\
\hline
\multirow{3}{*}{Blue Bottle} &  $M_{12}$ &  &  26.38 & 38.09\\
 & $M_{12}$  & $M_{15}$\tnote{a}  &  25.12 & 27.43\\

 & $M_{12}$ & $M_{16}$\tnote{b}  &  28.27 & 25.07\\

\hline 
\end{tabular}
\begin{tablenotes}
    \item[a] Model fine-tuned with robot practices only.
  \item[b] Model fine-tuned with robot practices plus human demonstrations.
  \end{tablenotes}
\end{threeparttable}
\caption{Accuracy of pouring water with the Blue Bottle fine-tuning model $M_{12}$.}
\label{tab-different-start-blue}
\end{table}

\subsection{Evaluation Summary}
Overall, the experiments and evaluation results show that 
\begin{enumerate}
\item The model trained on data of humans pouring water pours accurately using accustomed containers not seen during training. The mean volume error range of pouring water was from 4.12\,mL to 12.35\,mL for such containers.
\item The model also generalizes the accurate pouring behavior to liquids such as oil and syrup and solid materials such as rice and beans. 
\item By using robot practices as new training data, GSSP lowers the initial pouring error to values smaller than the state-of-the-art. For instance, experiments with a wine bottle showed a reduction of mean volume error from 51.22\,mL to 15.78\,mL (69\% reduction).
\item Batch or gradual fine-tuning yield similar results when there is enough variation on the practices collected.
\item Applying GSSP with the combination of human demonstrations and robot practices does not improve the accurate pouring generalization results.
\end{enumerate}

\section{Conclusion}
In this work, we presented a self-supervised learning from demonstrations approach that allows robots to pour as accurately and fast as humans. The presented work is based on a peephole LSTM that learns the motion dynamics by using the actual outcome of the demonstrations, regardless of an expert human execution. We evaluated the model using a robotic system that we devised and a UR5e robotic arm\footnote{Videos of the robotic pouring demos can be found at: \\
\texttt{https://youtu.be/MYfZBiHTDBc},\\ \texttt{https://youtu.be/u4OyQeMbwsQ}, and \\ \texttt{https://www.youtube.com/watch?v=xp9nEDTntU4}}. Based on the extensive experiments carried out, the presented model pours more accurately and faster than related works that have approached the accurate pouring problem. The capability of the model was further expanded with generalization by self-supervised practicing (GSSP) for containers and materials that presented high pouring error.
 
\section*{Acknowledgments}
This material is based upon work supported by the National Science Foundation under Grants Nos. 1812933 and 1910040.

\biboptions{square,comma}
\bibliography{root}

\end{document}